\documentclass[journal]{IEEEtran}
\usepackage{cite}
\usepackage{amsmath,amsfonts,amssymb,amsthm, mathtools}
    \DeclarePairedDelimiter{\norm}{\lVert}{\rVert} 
\usepackage{algorithmic}
\usepackage{mathtools}
\usepackage[]{graphicx} 
\usepackage{textcomp}
\usepackage{xcolor}
\usepackage[space]{grffile} 
\usepackage{float} 
\usepackage{array}  
\newcolumntype{P}[1]{>{\centering\arraybackslash}p{#1}}
\usepackage{titlesec}

\usepackage{xcolor}  

\usepackage[caption=false,font={small}]{subfig}
\usepackage[font=small,skip=2pt]{caption} 
\usepackage{csquotes}
\usepackage[space]{grffile}
\usepackage[]{cite} 

\usepackage{layouts}

\graphicspath{{figs/}}

\newcommand{\round}[1]{\ensuremath{\lfloor#1\rceil}}
\renewcommand{\tablename}{Table }

\begin{document}
\setlength{\abovedisplayskip}{3pt}
\setlength{\belowdisplayskip}{3pt}

\bstctlcite{IEEEexample:BSTcontrol} 

\title{Structural Prior Driven Regularized Deep Learning for Sonar Image Classification}

\makeatletter
\patchcmd{\@maketitle}
  {\addvspace{0.5\baselineskip}\egroup}
  {\addvspace{-1\baselineskip}\egroup}
  {}
  {}
\makeatother

\author{{Isaac~D.~Gerg and Vishal~Monga}
    \thanks{This work was partially supported by Office of Naval Research under grants N00014-19-1-2638, N00014-19-1-2513.
    
    I. D. Gerg is with the Applied Research Laboratory and School of EECS at the Pennsylvania State University. V. Monga is with the School of EECS at the Pennsylvania State University (http://signal.ee.psu.edu).}    
}

\maketitle


\begin{abstract}
Deep learning has been recently shown to improve performance in the domain of synthetic aperture sonar (SAS) image classification. Given the constant resolution with range of a SAS, it is no surprise that deep learning techniques perform so well.  Despite deep learning's recent success, there are still compelling open challenges in reducing the high false alarm rate and enabling success when training imagery is limited, which is a practical challenge that distinguishes the SAS classification problem from standard image classification set-ups where training imagery may be abundant.  We address these challenges by exploiting prior knowledge that humans use to grasp the scene. These include unconscious elimination of the image speckle and localization of objects in the scene. We introduce a new deep learning architecture which incorporates these priors with the goal of improving automatic target recognition (ATR) from SAS imagery.  Our proposal -- called SPDRDL, Structural Prior Driven Regularized Deep Learning -- incorporates the previously mentioned priors in a multi-task convolutional neural network (CNN) and requires no additional training data when compared to traditional SAS ATR methods. Two structural priors are enforced via regularization terms in the learning of the network: (1) structural similarity prior -- enhanced imagery (often through despeckling) aids human interpretation and is semantically similar to the original imagery and (2) structural scene context priors -- learned features ideally encapsulate target centering information; hence learning may be enhanced via a regularization that encourages fidelity against known ground truth target shifts (relative target position from scene center). Experiments on a challenging real-world dataset reveal that SPDRDL outperforms state-of-the-art deep learning and other competing methods for SAS image classification.
\end{abstract}



\section{Introduction}

Underwater sonar was historically pursued for military purposes, but as the field matured and the commercialization of these systems became feasible, remote-sensing applications for the civilian domain developed.  Predictably, as synthetic aperture sonar (SAS) was initially pursued for mine countermeasure applications \cite{stack2011automation}, it has broad civilian applications in remote-sensing of the undersea environment today.

\begin{figure}[!htbp]
    \begin{tabular}{c c c c}
        \includegraphics[width=0.2\linewidth,keepaspectratio]{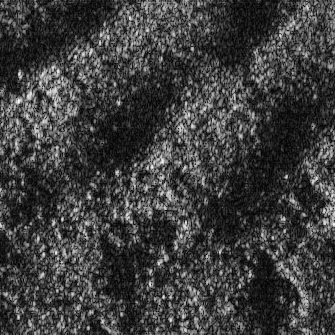} & 
        \includegraphics[width=0.2\linewidth,keepaspectratio]{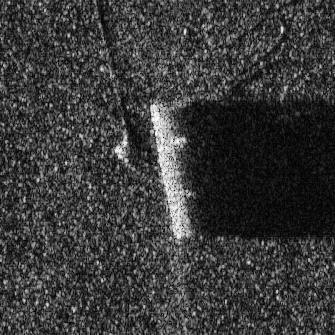} &
        \includegraphics[width=0.2\linewidth,keepaspectratio]{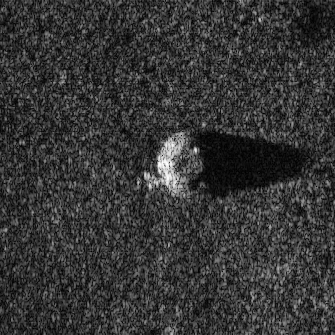} & 
        \includegraphics[width=0.2\linewidth,keepaspectratio]{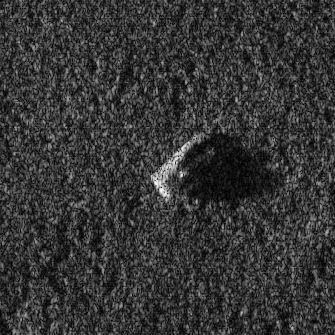}      
        \\
        (a) & (b) & (c) & (d) \\
        \includegraphics[width=0.2\linewidth,keepaspectratio]{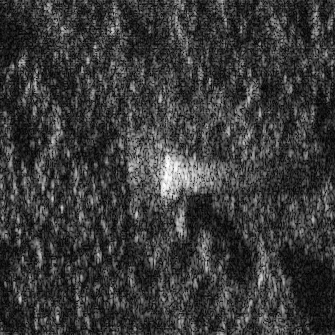}   &
        \includegraphics[width=0.2\linewidth,keepaspectratio]{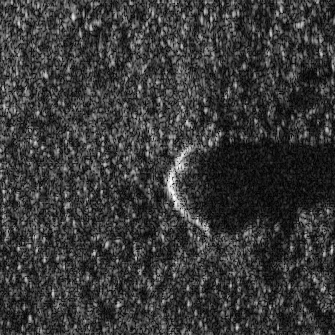}   &
        \includegraphics[width=0.2\linewidth,keepaspectratio]{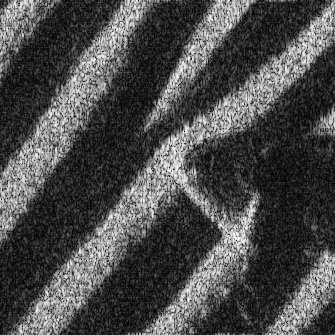}   &
        \includegraphics[width=0.2\linewidth,keepaspectratio]{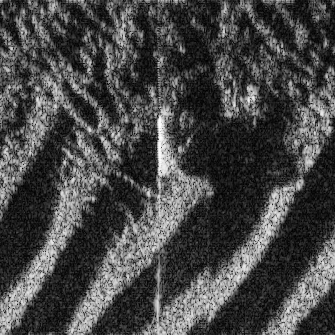}  \\    
        (e)  & (f)  & (g)  & (h)  
    \end{tabular}
    \caption{SAS is capable of producing high-quality, high-resolution imagery of seafloor and objects.  The images here are examples collected from the Centre for Maritime Research and Experimentation MUSCLE system. In this work, we provide an ATR algorithm to classify MUSCLE images into four classes.  Example objects from each class are shown: (a) background, (b) cylinder, (c) truncated cone, and (d) wedge. Difficulties in classification result because there are often objects which appear target-like (e,f), but are not targets (false alarms); and some targets are difficult to discern  (g,h) because of orientation, burial depth, or background topography causing them to be ignored (missed detections).}
    \label{fig:example_sas_images}
\end{figure}

 Over the last few years, SAS has matured to a capability accessible in the civilian space with several companies offering systems \cite{shea2014real} \cite{hansen2009synthetic}. 
 Fundamental work in obtaining high-quality SAS images was carried out in the late 1990's to early 2000's  \cite{gough1986synthetic, hayes1991results, hayes1992broad, callow2003signal, fortune2001statistical, hunter2003simulation}.
 Contemporary SAS systems are capable of producing image quality that lends itself to tasks such as automatic target recognition (ATR).  Despite the improvements, the problem of detecting and classifying objects in imagery remains challenging because of distractors in the environment and the complex configurations possible by targets.  \figurename \ref{fig:example_sas_images} shows examples of difficult cases along with prototypes of the object classes used in this work.

\subsection{Open Challenges in SAS ATR}

SAS ATR algorithms were originally established from ATR algorithms used in side scan sonar, which we call real aperture sonar (RAS).  RAS systems have similar collection geometry to SAS, but cannot produce the constant resolution with range achieved by SAS.  We refer the reader to Chapter Two of \cite{callow2003signal} and Chapter Three of \cite{hawkins1996synthetic} for the differences between RAS and SAS imaging. Despite these differences, initial SAS imagery looked similar enough to RAS for researchers to reasonably justify the use of RAS ATR algorithms on SAS.

Some of the more popular early work in sonar ATR involved the use of kernel filters \cite{hyland1995sea, dobeck1997automated}. When large amounts of SAS imagery began to be produced, these were some of the first techniques applied to it.  Over time, these methods began to utilize multiple target looks to aid in classification by exploiting the overlapped coverage of the seafloor most SAS surveys exhibit \cite{dobeck1999fusing}. Other techniques focused on model-based approaches \cite{reed2004automated} and uncertainty modeling. Eventually, the popular classification algorithms of the early 2000's, before the boon of deep learning, were applied to imagery including decision trees \cite{novakovic2009using} and Markov random fields \cite{reed2003automatic}.

Coincidentally, this paper is about the use of neural networks (NN) to address the classification problem. The use of such techniques is not new and one of the more popular early works employed them \cite{erkmen2008improving}.

With all the recent success of SAS, there remains persistent challenges with respect to ATR.  One of the biggest challenges for obtaining good results is collecting and labeling large amount of imagery which is needed for contemporary machine learning (ML) algorithms using deep learning.  SAS collection from unmanned underwater vehicles (UUV's) requires an inordinate amount of support infrastructure including: support vessels, ship crew, and divers making the endeavor financially expensive. Furthermore, the objects often sought upon during surveys are scarce.

It is also difficult to create an environment-independent ML algorithm when little training data is available.  Practitioners quickly discover that deploying classification in unseen environments often results in high false alarm rates, even for state-of-the-art SAS ATR algorithms.  Despite this, the detection performance of these methods is often quite good as the literature shows.  However, all the false alarms returned by the algorithm quickly overwhelm human operators.  Furthermore, objects simple to rule out by humans are often called by the ATR, preventing trust between the operators and the algorithms -- this renders the ATR useless.  When combined, these factors result in manual human inspection as a preferred means to cull the imagery; a costly process.

\subsection{Overview of Our Proposal}
We present a deep learning classifier exhibiting significantly reduced false alarm rates compared to contemporary SAS ATR algorithms while maintaining high detection accuracy. Our approach integrates high-level, domain knowledge unique to the SAS domain in order to achieve these good results.  We do this by integrating parcels of domain knowledge, which we call \emph{priors}, into the training objective.

\begin{figure}[t]    
    \centering
    \includegraphics[width=0.95\linewidth,keepaspectratio]{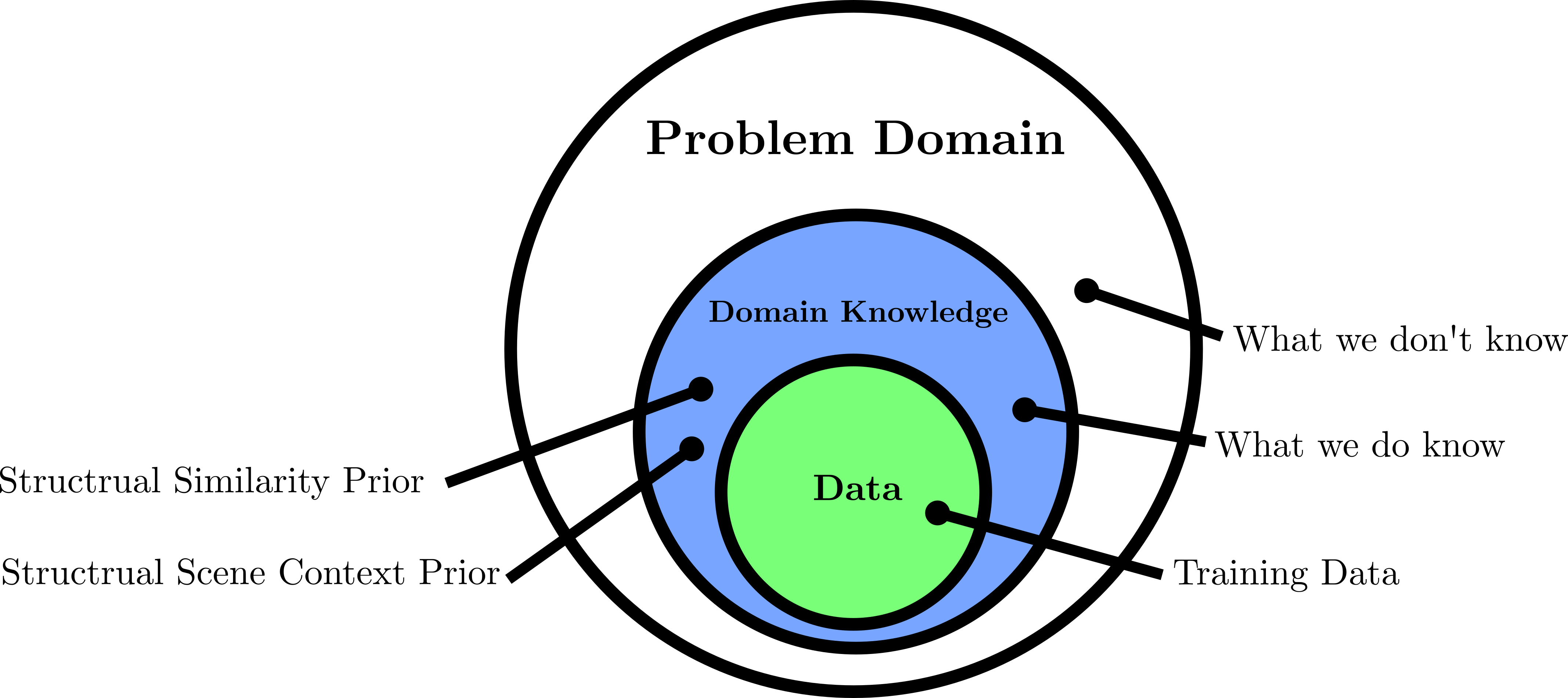}    
    \caption{The relationship of priors to the problem domain and the training data.  Our priors encapsulate domain knowledge which is not explicitly represented in the data but projected into the data.  The priors used in this work, structural similarity and structural scene context, are employed through regularization losses which are augmented to the primary task's objective function which is classification error.  We jointly train all losses so the network finds a minimum consistent with both the data and the domain priors.}
    \label{fig:venn_digram}
\end{figure}

For a given problem, there exist attributes which are directly represented by the given training data.  In the case of image classification, this would be the images/label pairs used for training; this information is explicitly provided to the training algorithm. This is the common scheme for the vast majority of image classification problems. However, there exists domain-specific information which is projected into the training data but may not be explicitly represented by it.  For example, for the dataset used in this work, we have some knowledge of how it was pre-processed before given to us for use.  Specifically, we know that the detection algorithm used to produce the image chips is generally good at centering targets within the chip. However, we do not have any kind of bounds or statistic on how well the targets are centered.  Furthermore, the true target centers are not explicitly encoded for each image.  Thus, the fact that the detector is reasonably good at centering the targets is domain knowledge derived from a subject matter expert (SME) (we will see this forms the scene context prior which will we discuss in future sections).  We refer to these parcels of domain knowledge as \emph{priors}.  In a deep learning framework, each prior is employed through a regularization loss which is augmented to the primary task's objective function. \figurename \ref{fig:venn_digram} is a Venn diagram illustrating the concept.

In this work, we define two priors which, when used individually each improve classification performance, but when used together, act synergistically to improve performance beyond the use of each exclusively.  The first prior we use addresses an often overlooked part of the SAS image reconstruction pipeline: image enhancement. This prior originates from the domain knowledge that image enhancement algorithms applied to SAS imagery aid in improving human interpretation.  An example of such an algorithm is despeckling \cite{poderico2012denoising}.  We name this prior the \emph{structural similarity prior} because it encapsulates the function of any image enhancement algorithm: improve scene content in a way which improves downstream task performance (in this case classification) while simultaneously preserving scene structure and semantics.  Quantitatively, this prior is captured by the regularization term in Eq (\ref{eqn:ms_ssim}) which is described in detail in Section \ref{section:SSP}.

The second prior we use leverages a common quality exhibited by detection algorithms: the ability to localize targets. The majority of SAS classification algorithms are preceded by a detection algorithm whose purpose is to quickly find target-like objects in the queue of mostly-benign seafloor images.  This prior originates from the domain knowledge that the detector algorithm is usually able to localize the target in the image which humans also do when parsing a scene.  We name this prior the \emph{structural scene context prior} because it encapsulates the role of  ground truth target position knowledge: well learned features for image interpretation should encode target location. The images output by the detector usually center the target and  we can translate the target through image crops. We then encourage prediction of the new ground truth target location but using the same features used for classification.  In this manner, we improve the quality of the features which consequently improves classification performance. Quantitatively, this prior is captured Eq (\ref{eqn:shift_invariance}) which is described in detail in Section \ref{section:SSCP}.

Recall that our two priors, structural similarity prior and structural scene context prior, exist for the purpose of improving our primary task: classification.  The domain knowledge captured by these two priors is employed through the use of regularization losses, Eq (\ref{eqn:ms_ssim}) and Eq (\ref{eqn:shift_invariance}),  augmented to the primary task objective function, Eq (\ref{eqn:categorical_focal_loss}).  Together this forms the final loss we jointly-optimize during training, Eq (\ref{eqn:ipsas_loss}).

Using the aforementioned priors above, \textbf{this paper makes the following technical contributions:
\begin{enumerate}
    \item Image  enhancement  through  despeckling  is  often  used to  improve  image  interpretability  for  humans.  We ask the question, \emph{Is there an image enhancement function which improves classification?} to which we will answer in the affirmative (results in \tablename \ref{table:ablation_losses}). To this end, we incorporate a data adaptive image enhancement network with a self-supervised, domain-specific loss to an existing classification network for purposes of improving classification performance. Our image enhancement function is learned from the  data removing  the  onerous  task  of  selecting  a fixed  despeckling  algorithm.  Furthermore,  ground  truth noise/denoised image pairs as required by previous methods \cite{vu2018deep} are not needed. 
    \item Most SAS ATRs only determine the presence of a target object but are aloof to where in the image it appears. To this end, we incorporate a target localization network in addition to a classification network for the purpose of also improving classification performance.  Like the image enhancement network, this is also trained using a self-supervised, domain-specific loss. Now, our classifier not only learns target class, but also target position thus acquiring scene context. Our target localization network is trained using the domain knowledge that objects are centered when passed to the classifier. We use the common data augmentation technique of image translation through cropping to induce new target positions when training, and have the target localization network estimate the induced target position in addition to the primary task of classification. This encourages the model to learn  ``where" of the scene in addition to the ``what".
    \item We train the two aforementioned networks and the classification network simultaneously through the addition of regularization terms to the primary classification loss objective function. This makes our formulation self-supervised and thus requires no extra data or labels making it suitable as a drop-in replacement for training against existing datasets. \tablename \ref{table:ablation_losses} shows through ablation that each domain-specific loss improves classification performance and that when combined, the best classification performance is achieved.
\end{enumerate}
Both of our priors incorporate structural domain knowledge so we call our method Structural Prior Driven Regularized Deep Learning (SPDRDL). Each prior mentioned has never used in previous SAS classification works. \tablename \ref{table:losses_and_types} shows the relationship among losses used in our final objective function.}

\begin{table*}[t]
    \caption{Brief summary of the loss terms in our objective function admitted from our domain priors. We list the relationships among the incorporated domain knowledge, the employed prior, and the associated regularization loss in our formulation.  We also list the primary task, classification, and its loss function for completeness.}
    \centering
    \begin{tabular}{c c c c}
        \hline
         \textbf{Prior Name} & \textbf{Employed Domain Knowledge} & \textbf{Loss Type}  & \textbf{Loss Equation} \\
        \hline \hline
        Structural similarity & Image enhancement improves human interpretability & Domain-specific  &  Equation \ref{eqn:ms_ssim} \\ 
        Structural scene context & Targets output from detector are image centered & Domain-Specific & Equation \ref{eqn:shift_invariance}\\
        N/A & N/A & Primary Task, Classification & Equation \ref{eqn:categorical_focal_loss} \\
        \hline
    \end{tabular}
    \label{table:losses_and_types}
\end{table*}

An overview of this paper follows. Section \ref{section:previous_work} provides a synopsis of past ATR approaches. Section \ref{section:proposed_method} presents the necessary background and development of SPDRDL and its use of domain priors. Section \ref{section:experiments} shows experimental results of our algorithm and compares our results to other contemporary algorithms on a challenging real-world dataset. Finally, Section \ref{section:conclusion} provides a summary of our findings.

\section{Previous Work} \label{section:previous_work}

Recent SAS ATR schemes have focused on improving feature representations through various means. For many years, representations were hand crafted and much of the research was in attempting to discover useful features through subject matter expert input. Techniques employed bag-of-words models \cite{Isaacs_2015_CVPR_Workshops} using handcrafted features as complex vocabularies describing SAS images.  Eventually, techniques emerged which removed the need for this explicit feature engineering task.  Dictionary learning methods were some of the first to forgo the explicit feature engineering path \cite{mckay2017robust, mckay2016localized} and automatically learn features as part of the classification process.  Today, deep learning techniques are employed in the same vein \cite{williams2016underwater}.

Recently, investigations into alternate representations to improve classification have shown be a fruitful endeavor.  \cite{gerg2018additional, williams2013benefit, williams2018exploiting} have examined representations derived from the $k$-space and have found they contain useful information for classification. Traditionally, the human consumable image, which arrives after extensive post-processing of the raw SAS data, has been used for input to the classifier.  The human consumable image is the result of a lengthy signal processing pipeline which discards information related to the frequency and direction of the received acoustic wavefronts.  A coarse explanation of a typical image reconstruction pipeline is as follows: (1) raw sonar echos are collected from the sonar array over multiple transmissions, (2) signal processing is applied to these echoes to correct them for imperfections, (3) the data is matched filtered to obtain resolution in the range dimension, (4) an optional motion compensation step is performed to interpolate the data to a regular grid (e.g. preparation for $\omega$-k beamforming), (5) the data is beamformed to generate a single look complex (SLC) image, and (6) a human consumable image is formed by taking the absolute value of the SLC and applying dynamic range compression (DRC). Consequently, the absolute value operation removes the phase portion of the SLC potentially discarding useful information.

Deep learning has been applied to sonar ATR resulting in a substantial improvement in classification performance. An initial work in the area is \cite{williams2016underwater} whereby convolutional neural networks (CNNs) were used to automatically learn features for classification. In \cite{mckay2017s}, the authors demonstrated the canonical transfer learning approach commonly used in training data-limited networks works well for SAS; a pre-trained CNN trained on the Imagenet dataset \cite{imagenet_cvpr09} was fine-tuned on SAS imagery yielding good results.  This work was generalized in \cite{mckay2018bridging}, where the authors integrate the feature learning of both SAS images and selected photographs simultaneously, yielding good ATR performance in the midst of limited training data. Finally, transfer learning among SAS sensors was demonstrated in \cite{williams2019transfer} where a CNN initially trained on one SAS sensor was used to quickly train with another.

In several of the works describe thus far, class imbalance has been mentioned as a noteworthy issue. Many SAS datasets have far more imagery of the benign seafloor than of objects of interest. To combat this issue, general adversarial network (GANs) have recently been applied to SAS for the purpose of generating more training data to balance the classes. In \cite{reed2019coupling}, a hybrid simulation and GAN based approach is used to generate a simulated, optical version of the desired scene and then a learned transform is applied to the simulated scene to give the appearance of a real SAS image. Their hybrid approach gives fine control over the generated scene content so the data balancing procedure can be accomplished with precision; particular objects, their orientations, and their range from the sonar can be specifically generated. Model based GAN approaches have not been limited to SAS, but also have been used for real-aperture sonar (RAS) systems as described in \cite{leedeep2019} where GANs are applied to RAS imagery to augment data for underwater person detection.

Today, SAS systems are multi-band and operate over several frequency ranges. This ability has not been overlooked in the context of ATR.  An early work utilizing multi-band sonars for classification is \cite{tucker2011coherence}. In this work, the authors demonstrate good detection performance when using a low-resolution broadband sonar in addition to a high frequency SAS.  Even more recently, deep learning has been applied to multi-band SAS imagery with good success and without the need of using a pre-trained network \cite{emigh2018supervised, galusha2019deep}. 

\section{Proposed Classification Method: SPDRDL} \label{section:proposed_method}

\subsection{Motivation of Approach}
Recent ATR schemes using deep learning demonstrate great performance but at the cost of requiring large amounts of training data.  As previously discussed, SAS data collection is costly resulting in small datasets which are almost always class imbalanced.  Consequently, it is crucial to use all the available information from a SAS image during classifier training.  To this end, we propose a new scheme which incorporates prior knowledge of SAS images in a novel way as a mechanism to extract more information from each image.  This additional information is used to positively influence classifier training.  

One mechanism by which we inject prior knowledge into the classification pipeline is by addressing the inherent speckle phenomenon present within every SAS image.  The speckle is often seen as noise and a nuisance for human interpretation. Much work has been done in the development of despeckling algorithms \cite{poderico2012denoising} with the purpose of enhancing image interpretability.  A natural outcome of this work is to ask if such types of enhancement are beneficial for improving classification performance and if so, which methods provide the most benefit.  Furthermore, can we forgo the onerous choice of selecting an enhancement algorithm algorithm and have the network learn the image enhancement transform in an unsupervised fashion?

Another prior which thus far has been overlooked in SAS ATR, are the assumptions given by the detector, sometimes called a pre-screener.  As background, traditional SAS ATR methods use a detector-classifier approach.  In this approach, a simple detection algorithm is first passed over the scene.  The detector produces candidate images of interest, sometimes called \emph{chips}, which are then passed to a classification algorithm for further inspection.  Usually the detector is computationally efficient and can quickly prune areas of the image which appear to be benign (e.g. a flat sandy sea-floor). Such a process reduces the amount of imagery the classifier has to process. It is believed such an approach was adopted initially for compute reasons -- early SAS classification systems were not capable of processing every possible sub-tile of an image in a timely manner due to limited compute power.  However, current compute capabilities, specifically in form of graphics processing units (GPUs), provide ample compute power enabling a classifier to examine whole scene quickly removing the need for the explicit detection step. Notwithstanding, for this approach to work, the classifier must be translation equivariant.

Good detectors can localize the target well and output SAS images with the target well-centered in the image.  The Mondrian detector \cite{williams2017mondrian} is a good example of such a detector. It uses prior knowledge of the target and sonar geometries to model expected relationships among local pixel neighborhoods; its quite capable for returning well-centered targets to a classifier.  However, current classifiers for SAS do not use this information.  They assume a target is present in the image, but do not explicitly estimate or assumes its position. On the other hand, our proposed method jointly estimates target class \emph{and} target position.

Because our proposed method estimates target position (in addition to object class), it is desirable to have a feature space embedding which is translation \emph{equivariant}. By equivariant, we mean that as the target translates smoothly across an image, its associated embedding also translates smoothly. Despite the convolutional nature of CNNs, they do not inherently provide translation equivariance.  Recent works such as \cite{zhang2016understanding, azulay2018deep, mairal2014convolutional} have pointed out this common misnomer and have made progress towards improvement. We utilize these techniques in our proposed method making it very robust to scene translation.  

Having the classifier robust to translations has an added benefit in that we can forgo the traditional detection step and run the classifier on across the entire scene.  This has an immediate benefit: the detection rate of the ATR is no longer bounded by the detector performance. For example, if a detector exhibits an eighty-percent detection rate, the overall ATR can do no better than an eighty-percent detection rate.  Hence, even with an oracle classifier, the best detection rate that can be achieved is eighty-percent.  

\subsection{Feature Extraction Network}
Traditional image classification pipelines using deep learning are composed of a feature extraction network followed by classification network.  Much recent work has been spent on designing an optimal feature extraction network as illustrated by the vast number of off-the-shelf (OTS) options available.  DensetNet \cite{huang2017densely}, Resnet \cite{he2016deep}, Inception \cite{szegedy2015going}, MobileNet \cite{howard2017mobilenets}, VGGNet \cite{Simonyan15}, and AlexNet \cite{krizhevsky2012imagenet} are popular examples of such OTS networks.  We leverage the good results these OTS network architectures and begin the construction of SPDRDL around a popular one: DenseNet-121.  SPDRDL is composed of Densetnet-121 as a feature extraction network (pink box) followed by a standard classification network (yellow box) shown in \figurename \ref{fig:network}.


\subsection{Structural Similarity Prior Via Data-Adaptive Image Enhancement Network} \label{section:SSP}
Building upon the feature extraction and classifier networks, we introduce a data-adaptive image enhancement network which is added to the front of the feature extraction network.  This enhancement network is shown in the blue box in \figurename \ref{fig:network}. The purpose of this network is to learn an image transformation which improves classification performance while still maintaining the original image semantics by obtaining an enhanced image (i.e. enhanced for classification purposes not necessarily human consumption) that is structurally similar to the original. We implement this network as a U-Net architecture \cite{ronneberger2015u} with the original image as input and the enhanced image as output.

To encourage image enhancement for classification, we utilize a novel loss function between the desired enhanced image and the original, the multiscale structural similarity measure (MS-SSIM) \cite{wang2004image, wang2003multiscale} which is a scale-aware version of the SSIM measure,
\begin{multline}
    \text{SSIM}(\mathbf{x}, \mathbf{y}) = {\underbrace{\left( \frac{2\mu_x \mu_y + C_1}{\mu_x^2 + \mu_y^2 + C_1} \right)}_\text{luminance}}^{\alpha} \cdot {\underbrace{\left( \frac{2\sigma_x\sigma_y + C_2}{\sigma_x^2 \sigma_y^2 + C2} \right)}_\text{contrast}}^\beta \\
    \cdot {\underbrace{ \left( \frac{\sigma_{xy}+C_3}{\sigma_x \sigma_y + C_3} \right)}_\text{structural}}^\gamma
    \label{eqn:ssim}
\end{multline}
where $\mathbf{x}$ and $\mathbf{y}$ are images being compared, $\mu$ and $\sigma$ are the patch-wise mean and standard deviation respectively of the corresponding image, $\sigma_{xy}$ is the covariance between image $\mathbf{x}$ and $\mathbf{y}$, $\{\alpha, \beta, \gamma\}$ are shaping constants, and $C_{1,2,3}$ are calibration constants. $\text{SSIM}(\mathbf{x}, \mathbf{y})  \in [0,1]$ where higher values indicate higher perpetual similarity between the images. The SSIM is differentiable and tractable for incorporation into a deep learning network \cite{tofighi2019prior}. 

MS-SSIM introduces scale dependence by computing structural and contrast factors of SSIM over several staged, low-pass-filtered versions of the input image and then combining their results. It is given by,
\begin{equation}
    \text{MS-SSIM}(\mathbf{x}, \mathbf{y}) = [l_M(\mathbf{x}, \mathbf{y})]^{\alpha_M} \prod_{j=1}^M[c_j(\mathbf{x}, \mathbf{y})]^{\beta_j}[s_j(\mathbf{x}, \mathbf{y})]^{\gamma_j}
    \label{eqn:msssim}
\end{equation}
where functions $l, c$ and $s$ represents the corresponding luminance, contrast, and structural components of Eq (\ref{eqn:ssim}) respectively, and $M$ is the total number of scales to evaluate. For all constants, we use the same values as specified in \cite{wang2003multiscale}.

By seeking to maximize the MS-SSIM between the original and the enhanced image in the enhancement net, we leverage human visual system priors designed into the MS-SSIM perceptual loss function \cite{wang2009mean}.   This utilizes the desired domain knowledge we which seek to embed in our formulation: there exists an enhanced imaged which is structurally similar to the original image but is able to yield improved classification performance.
Finally, we define the structural similarity prior (SSP) regularization loss as,
\begin{equation}
    \mathcal{L}_{\text{SSP}}(\mathbf{x}, \mathbf{x}_{\text{enhanced}}) = 1 - \text{MS-SSIM}(\mathbf{x}, \mathbf{x}_{\text{enhanced}})
    \label{eqn:ms_ssim}
\end{equation}
where $\mathbf{x}$ is the input image and $\mathbf{x}_{\text{enhanced}}$ is the improved image output by the enhancement network. Without this loss term, the network has no notion of a ``noise model" and simply seeks to minimize the weights of the function with no understanding of the hand-crafted network structure we designed to exploit the domain prior.

\subsection{Structural Scene Context Prior Via Target Localization Network} \label{section:SSCP}

As previously mentioned, the detector returns targets centered in the image. During the data augmentation process, these targets are translated by a random amount.  This augmentation procedure is commonly used in other SAS ATR methods.  However, our method is different in that we do not discard the translation parameters but encourage the network to estimate them while simultaneously performing classification.  In this manner, we embed the target position domain knowledge into the feature extraction network by encouraging it to learn a spatially-aware context of the scene in addition to features for classification.  With this prior, the likelihood of the network learning features which are not target-centric is reduced, and the creation of features derived from biases within the dataset, like seafloor texture, is reduced.

\begin{figure*}[t]
    \centering
    \includegraphics[width=0.999\linewidth]{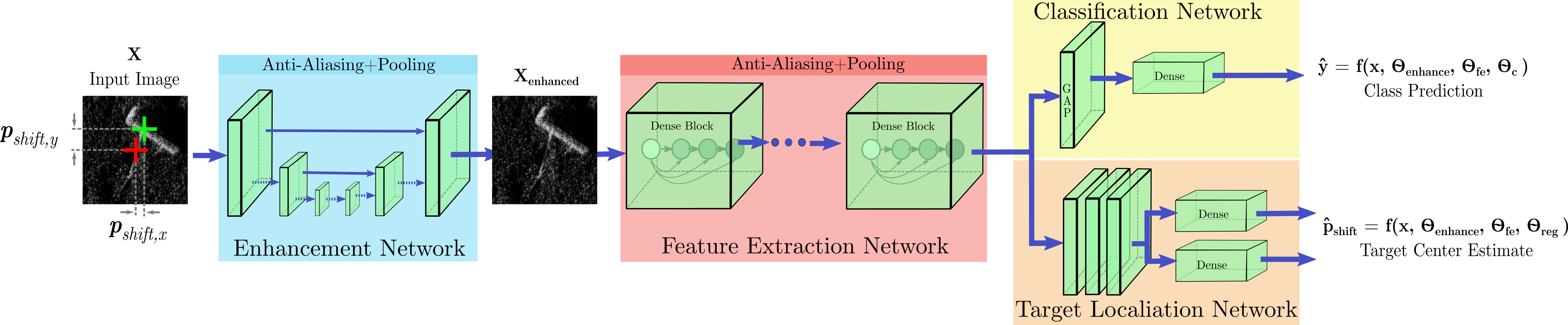}
    \caption{The SPDRDL network architecture; the network input is a SAS image and the output is a classification and target center position when a target is present. SPDRDL is composed of four modules: image enhancement network, feature extraction network, target localization network, and a classification network. SPDRDL leverages two domain priors to improve classification: (1) image enhancement algorithms like despeckling improve image interpretability and (2), the detector produces SAS images with well-centers targets. For the former, an enhancement network leverages the human visual system priors incorporated into the structural similarity prior to enhance the image for classification. For the latter, input images are translated as part of the data augmentation procedure during training and this translation is estimated in addition to predicting image class. Image classification, enhancement, and target localization are simultaneously trained.}
    \label{fig:network}
\end{figure*}

We encourage the network to learn target localization by augmenting our feature extractor with a target localization network whose task is to estimate the target position from the feature embedding. Recall that the ground truth for this estimate is determined through the data augmentation procedure.  The target localization network is represented by the orange box in \figurename \ref{fig:network}. It is composed of a set of 1 $\times$ 1 convolutions to reduce the dimensionality of the embedding. This reduction serves as a bottleneck which then feeds two dense layers which both have no post-activation function simply returning the position estimates.  Formally, we define the structural scene context prior (SSCP)  regularization loss as,
\begin{equation}
    \mathcal{L}_{SSCP}(\mathbf{p}_{shift}, \hat{\mathbf{p}}_{shift}) =  \frac{1}{2}\norm{\mathbf{p}_{shift} - \hat{\mathbf{p}}_{shift}}^2
    \label{eqn:shift_invariance}
\end{equation}
where $\mathcal{L}_{SSCP}$ represents the mean-squared error between the shift (i.e. translation) applied during data augmentation, $\mathbf{p}_{\text{shift}}$, and the shift estimated by the network, $\hat{\mathbf{p}}_{\text{shift}}$.

\begin{figure*}[t]
    \centering
    \includegraphics[width=0.75\linewidth]{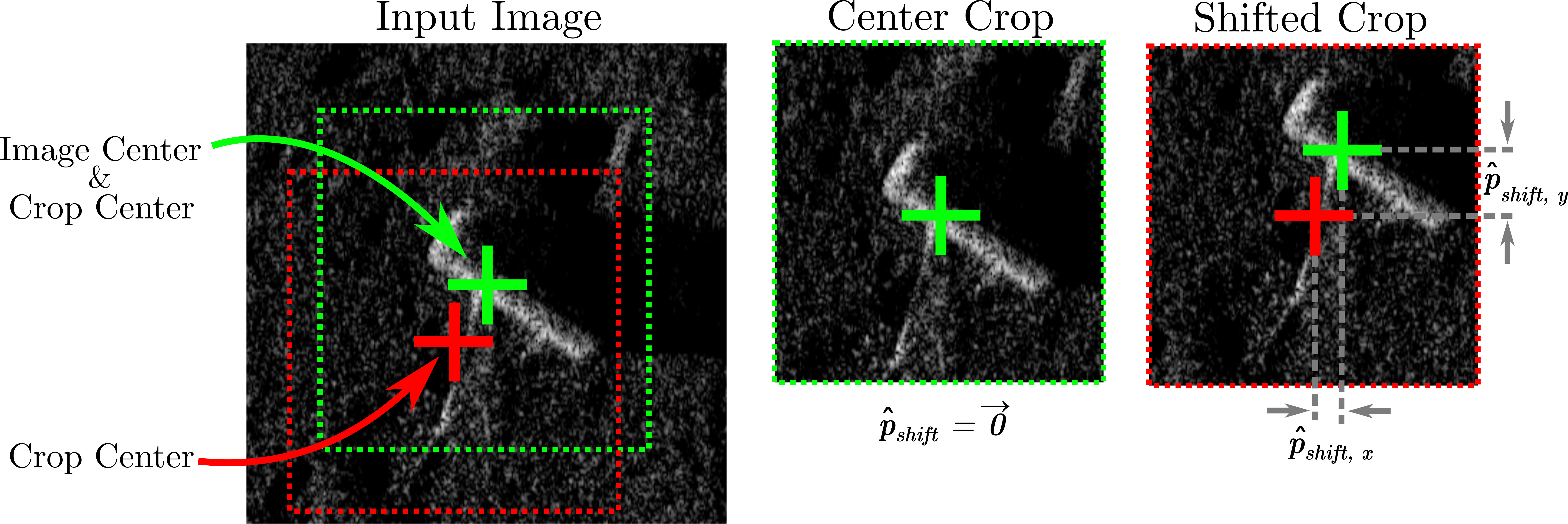}
    \caption{The detector returns well-centered target images.  We can use this prior during our data augmentation procedure.  During data augmentation, we translate the image via random crops and record the effective translation shift induced as $\mathbf{p}_{\text{shift}}$.  To incorporate this prior into the network, we add a target localization task which encourages the network to estimate this translational shift in addition to outputting classification. As an example of this procedure, the detector returns a well-centered target image (left) and random crops of this image are fed to the network as data augmentation (middle, right).  The target localization network estimates the difference between the true center of the target (green) and the shifted center (red). The translation shift estimate is denoted as $\mathbf{\hat{p}}_{\text{shift}}$.}
    \label{fig:pshift}
\end{figure*} 

\figurename \ref{fig:pshift} shows an example of how the positional shifts are created.  First, the Mondrian detector returns an image with the target centered. Next, a random crop is applied to the image during data augmentation.  This induces a translation of the target in the image.  We denote this translation as ${\mathbf{p}}_{\text{shift}}$ and add this information to the network via the backpropagation through the loss of Eq (\ref{eqn:shift_invariance}).

Thus far in this sub-section, we have developed a novel method to encourage the network to learn target position within the scene, and we have done so in a self-supervised fashion.  One assumption we have made which have not yet addressed is to assume that the feature space is translation equivariant.  Despite the use of convolutional layers in our network, non-unitary strides associated with convolution and pooling operations prevent translation equivariance as we will show. Additionally, we have not provided local pixel positioning information to the network likely resulting in position information being determined by specific neurons in the dense layers which is undesirable for generalization.  In the next two subsections, we will address each assumption.

\subsubsection{Addition of Anti-Aliasing Filtering Before Pooling Layers}
Most CNNs are not inherently shift invariant \cite{azulay2018deep} when combined with pooling layers.  This is caused by the lack of proper filtering done during image subsampling in pooling  and convolutional layers when the stride is greater than one. Strided layers perform two operations: (1) a filtering procedure which is run over the entire image (in the case of max pooling, this is an order-statistic filter), and (2), image subsampling to reduce the image dimensions most commonly done by striding to reduce the compute burden.  The striding operation is subsampling the image for the purposes of decimation.  During this procedure, the energy of the discarded frequencies is folded into the desired lower frequency band reducing the signal-to-noise ratio (SNR) of the resulting embedding.  This results in a feature space which is not translation \emph{equivariant}: translations in the input image do not correspond to translations in the feature embedding.  

We can overcome the faults of the traditional pooling layer by introducing an anti-aliasing (AA) filter before all strided operations \cite{zhang2019making}. In our setup, this means placing an AA filter before all strided convolutions and pooling operations. The AA filters prevents out-of-band frequencies from aliasing back into the remaining spectrum post-subsampling.  This results in increasing the signal-to-noise ratio (SNR) of the embedding and to encourage translation \emph{equivariance}. 


\subsubsection{Feature Position Encoding}
As previously mentioned, CNNs do not naturally provide translation invariance when used in tandem with strided convolution and/or pooling layers.  In addition,  \cite{liu2018intriguing} demonstrated that CNN's have difficulty with position oriented tasks because they do not encode feature position.  Indeed, this at first seems surprising given the translational nature of the convolution operator. However, the convolution operator takes as input a 2D map and also outputs a 2D map; a feature's position through this process is a function of the map domain but not explicitly coded in the representation.  Hence, when a 2D map is flattened and used as input to a dense layer, a feature's position is lost.

The interesting problem of CNNs not recording positional information was not only noted by \cite{liu2018intriguing}. In an early and popular work, \cite{zhao2015stacked} noted that CNNs are good at providing  ``what" but not the ``where."  They specifically design their CNN architecture to compensate for this fault.  Furthermore, the supplementary material of \cite{ulyanov2018deep} also notes this as they cite the addition of positional information improved image in-painting tasks for their Deep Image Prior technique.  

We augment SPDRDL with target position information by using the CoordConv solution of \cite{liu2018intriguing}.  In particular, we augment the output of the image enhancement network, $\mathbf{x}_{\text{enhanced}}$, with two additional channels, each one describing a positional dimension of the input map as described by Eq (\ref{eqn:coordconv2}),
\begin{equation}
\label{eqn:coordconv2}
    \begin{split}
        \mathbf{V}, \mathbf{H} \in \mathbb{R}^{ h \times w} \\
        \mathbf{V} [k, l] = l/w - \round{l/2} \\
        \mathbf{H} [k, l] = k/h - \round{h/2}
    \end{split}
\end{equation}
where $\mathbf{V}$ and $\mathbf{H}$ are the generated 2D maps augmented to the channel dimension of the input, $h, w$ are the height and width of the image respectively in pixels, and $k, l$ are pixel locations.


\subsection{Classification Loss}
Categorical cross-entropy is a commonly used loss function for penalizing classification error in neural networks.  It accounts for errors probabilistically by measuring the amount of surprise between the predicted and true labels. The measure works well in the presence of balanced class and accurate labels.  However, we know for SAS that the number of negative examples far outweighs the positive examples.

To mitigate the shortcomings of categorical cross-entropy in the presence of class imbalanced, we use a specified weighted version of the measure called the focal loss \cite{lin2017focal}. The focal loss is given by Eq (\ref{eqn:categorical_focal_loss}),
\begin{equation}
\mathcal{L}_{\text{FL}}(\mathbf{y}, \mathbf{\hat{y}}) = \sum_{c=1}^N -\alpha (1-{\hat{y}}_c)^\gamma y_c \log(\hat{y}_c)
\label{eqn:categorical_focal_loss}
\end{equation}
where $N$ is the number of classes, $y_c$ is the true probability of class $c$, $\hat{y}_c$ is the estimated probability of class $c$, and we use the strength coefficients given by the paper of $\alpha=0.25$ and $\gamma = 2$. Focal loss is a weighted version of the cross-entropy loss whereby correct classification is de-weighted.  Consequently, the effect of the focal loss is to place more emphasis on grossly mis-classified samples compared to virtually correct classified samples.  Through the use of the focal loss, error gradients of correct classifications are greatly diminished during training time while grossly incorrect classification maintain their error magnitude. In this way, the focal loss focuses the training on the misclassification samples and largely leaves the easy, correct classifications untouched.

There are several ways to place emphasis on negative samples during training of which a common one is to assign label weights.  However, we chose the focal loss because of several positive properties it offers for our setup.  Following, we describe each.

The first benefit realized by focal loss is that it can be viewed through the lens of importance sampling \cite{katharopoulos2018not} but without the explicit overhead associated with such techniques. Recently, \cite{williams2013benefit} showed importance sampling works well to improve the performance of SAS ATR. In importance sampling, misclassified samples are shown more often during the training procedure than correctly classified samples.  In a similar manner, using the focal loss can instill a similar training policy without the overhead of maintaining a list of the misclassified samples. Using the focal loss, a mini-batch of images is fed to the training algorithm and the misclassified samples are dynamically weighted proportional to their error.  For each batch, correctly classified samples induce little error gradient and effectively are removed from the batch.

The second benefit realized by focal loss is that the effective batch size is reduced over time. Reducing batch sizes has been associated with better generalization error \cite{keskar2017}. Assuming the distribution of easy- and hard-to-classify samples is uniform throughout the minibatch,  at the beginning of training all samples in the minbatch are considered hard-to-classify.  As training progresses, some samples become easier-to-classify and their error gradients vanish effectively removing them from the minibatch reducing the effective minibatch size.

\subsection{SPDRDL: Jointly Learned Image Enhancement and Object Location Estimation}
In the previous sections, we examined sources of structural information in SAS images currently not utilized by contemporary ATR methods. Our proposed approach builds upon an existing CNN backbone network commonly used for feature extraction by utilizing this overlooked structural information. In this section, we bring together the aforementioned sections and fully present our proposed method, SPDRDL.

Incorporating the losses discussed in the previous section, we arrive at the final loss function for SPDRDL, Eq (\ref{eqn:ipsas_loss}),
\begin{equation}
    \begin{aligned}
        \mathcal{L}(\mathbf{\Theta}_{\text{enhance}}, & \mathbf{\Theta}_{\text{reg}}, \mathbf{\Theta}_{\text{c}}, \mathbf{\Theta}_{\text{fe}}) =
        \\ &\mathcal{L}_{\text{FL}}(\mathbf{y}, \mathbf{\hat{y}}, \mathbf{\Theta}_{\text{c}}) \\ 
        & + \lambda_1   \mathcal{L}_{\text{SSP}}(\mathbf{x}, \mathbf{x}_{enhanced}, \mathbf{\Theta}_{\text{enhance}}) \\ 
        & +  w \lambda_2 \mathcal{L}_{SSCP}(\mathbf{p}_{shift}, \hat{\mathbf{p}}_{shift}, \mathbf{\Theta}_{\text{reg}})
    \end{aligned}
    \label{eqn:ipsas_loss}
\end{equation}
where $\mathbf{x}$ is the input image, $\mathbf{x}_{\text{enhanced}}$ is the data adaptive enhanced image, $\mathbf{y}$ is the true target class, $\mathbf{\hat{y}}$ is the predicted target class, $\mathbf{p}_{shift}$ is the true target translation, $\hat{\mathbf{p}}_{shift}$ is the estimated target translation,  $\mathbf{\Theta}_{\text{reg}}$ are the target localization network parameters,  $\mathbf{\Theta}_{\text{enhance}}$ are the image enhancement network parameters, $\mathbf{\Theta}_{\text{c}}$ is the classification network parameters, $\mathbf{\Theta}_{\text{fe}}$ is the feature extraction network parameters, and $\lambda_1, \lambda_2$ are regularization weights. Finally, $w$ is a class-dependent weight for the localization task given by Eq (\ref{eqn:weight}),
\begin{equation}
    w = \begin{cases}
    0 &\text{background class}\\
    1 &\text{any target class}
    \end{cases}
    \label{eqn:weight}
\end{equation}

SPDRDL's network description is in \tablename \ref{table:network_arch}. Convolutional layers are followed by ReLU activation and use initialization of \cite{he2015delving}.  Anywhere subsampling was used (which includes pooling layers and strided convolutions), anti-aliasing filtering was applied before subsampling using a 3 $\times$ 3 kernel of $\frac{1}{16}\begin{bmatrix}
1 & 2 & 1\\
2 & 4 & 2 \\
1 & 2 & 1
\end{bmatrix}$.

\begin{table}[t]
   \caption{Description of SPDRDL architecture network architecture. The network input is a 256 $\times$ 256 pixel grayscale SAS image normalized to $[0,1]$. The network has two outputs, a  classification output and a target position output. AA indicates anti-aliasing was applied to the layer.}
    \centering
    \scalebox{0.8}{
        \begin{tabular}{c c c c c}
            \hline
            \textbf{Layer Name} & \textbf{Layer Function} & \textbf{Dimensions} & \textbf{\# Filters} & \textbf{Input}\\ 
            \hline\hline
            input1 & Input & N/A & N/A & N/A \\  
            conv1a & Convolution & 3x3 & 16 & input1 \\
            conv1b & Convolution & 3x3 & 16 & conv1a \\
            pool1 & AA Max Pooling & 2x2 & N/A & conv1b \\
            \hline
            conv2a & Convolution & 3x3 & 32 & pool1 \\
            conv2b & Convolution & 3x3 & 32 & conv2a \\
            pool2 & AA Max Pooling & 2x2 & N/A & conv2b \\
            \hline
            conv3a & Convolution & 3x3 & 64 & pool2 \\
            conv3b & Convolution & 3x3 & 64 & conv3a \\
            pool3 & AA Max Pooling & 2x2 & N/A & conv3b \\
            \hline
            conv4a & Convolution & 3x3 & 128 & pool3 \\
            conv4b & Convolution & 3x3 & 128 & conv4a \\
            pool4 & AA Max Pooling  & 2x2 & N/A & conv4b \\
            \hline
            conv5a & Convolution & 3x3 & 256 & pool4 \\
            conv5b & Convolution & 3x3 & 256 & conv5a \\
            \hline
            up1 & Upsampling & 2x2 & N/A & conv5b \\
            conv6a & Convolution & 3x3 & 128 & up1 \\
            merge1 & Concatenate & N/A & N/A & conv6a, conv4b  \\
            conv6b & Convolution & 3x3 & 128 & merge1 \\
            conv6c & Convolution & 3x3 & 128 & conv6b \\
            \hline
            up2 & Upsampling & 2x2 & N/A & conv6c \\
            conv7a & Convolution & 3x3 & 64 & up2 \\
            merge2 & Concatenate & N/A & N/A & conv7a, conv3b  \\
            conv7b & Convolution & 3x3 & 64 & merge2 \\
            conv7c & Convolution & 3x3 & 64 & conv7b \\
            \hline
            up3 & Upsampling & 2x2 & N/A & conv7c \\
            conv8a & Convolution & 3x3 & 32 & up3 \\
            merge3 & Concatenate & N/A & N/A & conv8a, conv2b  \\
            conv8b & Convolution & 3x3 & 32 & merge3 \\
            conv8c & Convolution & 3x3 & 32 & conv8b \\
            \hline
            up4 & Upsampling & 2x2 & N/A & conv8c \\
            conv9a & Convolution & 3x3 & 16 & up4 \\
            merge4 & Concatenate & N/A & N/A & conv9a, conv1b \\
            conv9b & Convolution & 3x3 & 16 & merge4 \\
            conv9c & Convolution & 3x3 & 16 & conv9b \\
            \hline
            conv9d & Convolution & 3x3 & 2 & conv9c \\
            conv9e & Convolution & 1x1 & 1 & conv9d \\
            lambda1 & $\frac{in - min(in)}{max(in)-min(in) + \epsilon}$ & N/A & N/A & conv9e \\
            \hline
            densenet1 & AA Densenet121 & N/A & N/A & lambda1 \\
            \hline            
            gap1 & Global Average Pooling & N/A & N/A & densenet1 \\
            classification & Dense with softmax & 4 & N/A & gap1 \\
            \hline   
            conv10 & Convolution & 1x1 & 256 & densenet1 \\  
            conv11 & Convolution & 1x1 & 128 & conv10 \\ 
            conv12 & Convolution & 1x1 & 64 & conv11 \\     
            flatten1 & Flatten layer & N/A & N/A & conv12 \\
            xPosEstimate & Dense  & 1 & N/A & flatten1 \\
            yPosEstimate & Dense  & 1 & N/A & flatten1 \\                        
            \hline 
    \end{tabular}}
    \label{table:network_arch}
\end{table}

\section{Experiments} \label{section:experiments}
In this section, we describe how we measure the performance of SPDRDL and demonstrates its efficacy against contemporary methods.  First, we will describe how we setup the experiments.  Next, we describe the comparison methods. Finally, we show results by comparing all the methods.

\subsection{Setup}
The ultimate goal of our experiments is to show the superiority of SPDRDL over existing methods. Equally important are two regimes to characterize.  The first regime is classification performance of each object class.  We show results in this regime by using confusion matrices whose purpose is to provide an overview of the classifier accuracy as a function of class.  The second regime is classification performance in a one-versus-all scenario, whereby the target classes are collated into a single group.  This collation converts our four class problem to a two class problem consisting of a target class and background class.  Additionally, we will use a variation of this regime by showing performance of a particular target class versus all others.  

As mentioned, we present results in a one-versus-all regime through conversion of a multi-class problem into a binary one.  For binary class problems, many metrics exist by which to measure efficiency.  A popular method is to measure area under the receiver operating characteristic curve (AUCROC). AUCROC reports the statistics of any chosen pair of samples being classified correctly.  However, the method has been shown to be sensitive to class imbalance \cite{davis2006relationship} which is pervasive here.  Therefore, we choose  area under the precision-recall (AUCPR) as our performance metric based on the analysis of \cite{davis2006relationship} which determined that AUCPR is superior over AUCROC when the number of negative class samples greatly outnumbers the positive class samples which is true here. Furthermore, \cite{davis2006relationship} demonstrate that the stability of AUCPR over AUCROC meaning a performance curve dominating in precision-recall (PR) space also dominates in ROC space but not vice-versa.

As previously mentioned, we will use confusion matrices to measure per class accuracy.  For the one-versus-all cases mentioned, we will use AUCPR.  This metric has a benefit over a confusion matrix because it does not force us to specify a threshold as all thresholds are evaluated. Usually, a threshold is set to optimize for a specific performance metric which is context dependent.  In lieu of having to select a particular context, we simply report AUCPR on a one-versus-all basis.  

Sonar image fidelity is often a function of range.  For example, spreading and absorption losses in the medium attenuate the reflected signals as a function of range.  Therefore, the SNR of sonar echoes is reduced at long ranges.  To measure the contribution of such effects on our classifier, we evaluate the classifier performance as a function of observation range.  

Recalling that CNNs with strides or pooling are not translation invariant, we also evaluate translation performance. Ideally, translation invariance would be evaluated at every possible translation of the target but this becomes prohibitively expensive to compute.  So, we evaluate translation invariance at eight extreme shifts of 59cm as shown in \figurename \ref{fig:image_augmentation}. Good translation invariance will yield the same classification regardless of shift so we compute performance by measuring the standard deviation (stdev) of the output score as a function of these nine shifts (eight extreme shifts plus the center crop).

\subsection{Dataset Description}
We train and evaluate SPDRDL on the dataset of images which are output from the Mondrian detector of SAS imagery collected by the CMRE MUSCLE SAS sensor \cite{baralli2013gpu}.  It is the same dataset used in \cite{gerg2018additional} but with three modifications: (1)
The original dataset contains detections on image boundaries and these images are extrapolated by mirroring resulting in target shapes which are not seen in the real environment. (2) Some of the images contained quadratic phase error (QPE) based on visual inspection \cite{cook2008analysis}. We removed this error by applying a brute force autofocus in the $k$-space domain. (3) The images were dynamic range compressed using an algorithm based on the rational mapping function of \cite{schlick1995quantization}.     

\begin{figure}[t]
    \centering
    \begin{tabular}{c}
        \centering
        \includegraphics[width=0.95\linewidth]{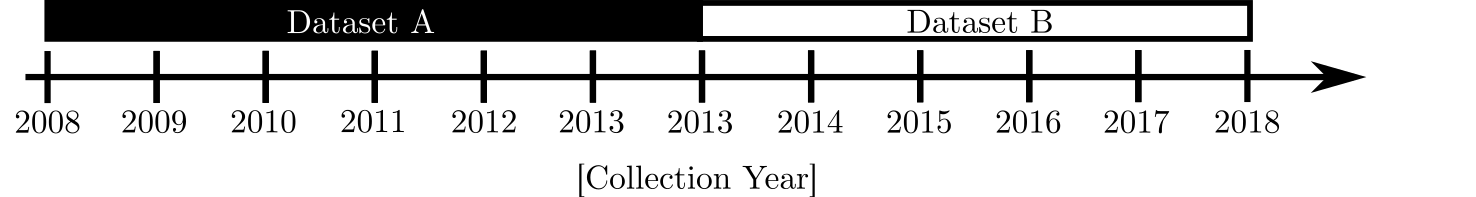} 
    \end{tabular}
    \caption{The dataset we use for our experiment is split into two groups based on collection year.  Dataset A is used for the training set and Dataset B is used for the validation and tests sets.}
    \label{fig:train_test_years}
\end{figure}  

Overall, the dataset is composed of two partitions (dataset A and dataset B) based on collection year which \figurename \ref{fig:train_test_years} depicts. Dataset A is composed of 27,748 images containing 1,385 targets collected from 2008 through 2013.  Dataset B from a set of 21,181 images composed of 639 targets collected from 2013 through 2018. Each image in the dataset has resolution of 1.5cm and of size 335 $\times$ 335 pixels. However, translation is induced through cropping the images down to 256 $\times$ 256 pixels at training/inference time.

\begin{figure}[t]
    \begin{tabular}{c c}
        \includegraphics[width=0.45\linewidth]{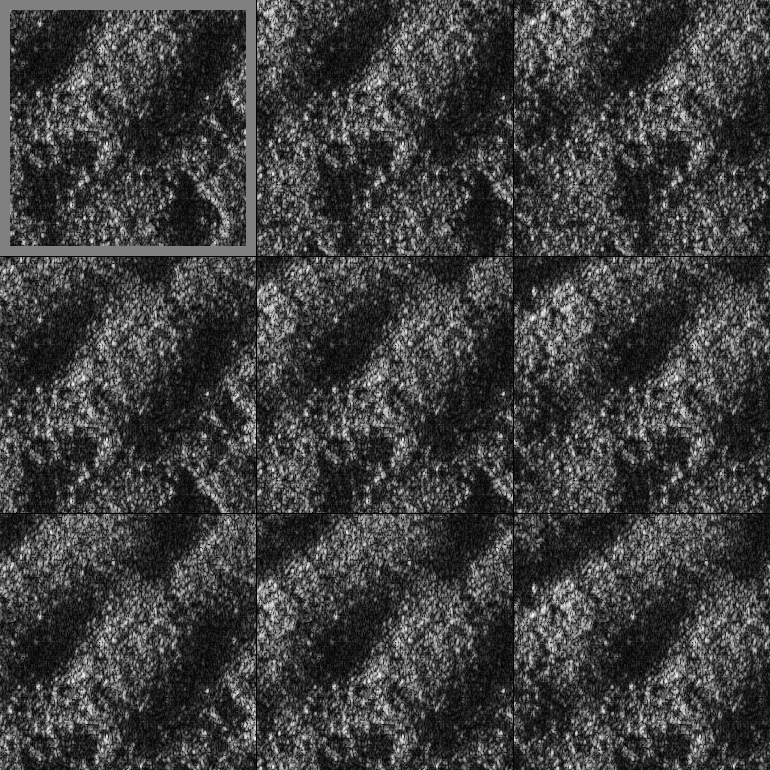} &
        \includegraphics[width=0.45\linewidth]{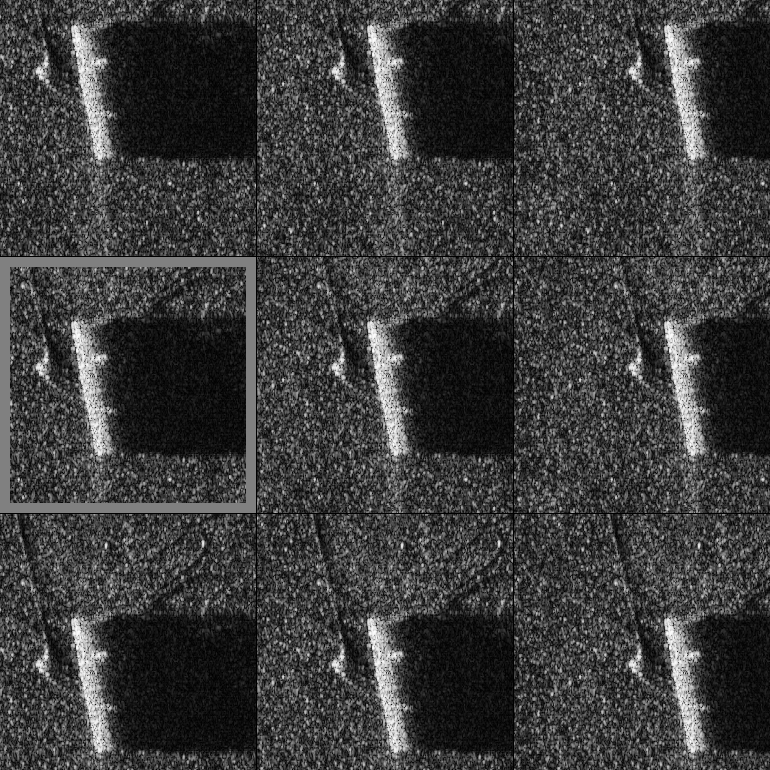} \\
        (a) & (b)       
    \end{tabular}
    \caption{The validation and test sets are derived form eight-way neighbor translations of 59cm.  The original tile is from Dataset B and cropped. All nine croppings are used for the test set with a random crop used for the validation set shown by the boxed imaged.  We show the test and validation data generation scheme for example images of the (a) background class and (b) target class.}
    \label{fig:image_augmentation}
\end{figure}  

The data is partitioned into three sections for evaluation purposes:
(1) The training set is composed of Dataset A and augmentations of it. (2) The validation set is composed of a single translation of each image of Dataset B. (3) The test set is composed of nine translations of each image of Dataset B.

The translations applied to the validation and test set are from the set of shifts in the horizontal and vertical dimensions of the set $\{-59\text{cm}, 0\text{cm}, 59\text{cm}\}$.  This configuration yields a total of nine possible shifts (including the center crop which is not shifted at all) for each image of the test set.  The validation set is composed of one random crop from the set of nine available for each image.  \figurename \ref{fig:image_augmentation} shows two examples of how the proposed cropping scheme contributes to the test and validation sets.  In the examples, a given image from Dataset B is eight-way shifted 59cm yielding a total of nine images (the mosaic) which are assigned to the test set and of one image (bounded by the solid line) which is randomly selected to be in the validation set. Overall, the training set is composed of 27,748 images, the validation set is composed of 21,181 images, and the test set is composed of 190,629 images.

\subsection{SPDRDL Training Procedure}
We train SPDRDL in a similar fashion to most other CNNs.  We use a mini-batch size of sixteen of which we assign half the batch an image from the background class and half the batch an image from the one of the three target classes. This 50:50 background-to-target-class split is based on the analysis of \cite{wallace2011class}.  Recall, images from the dataset are 335 $\times$ 335 pixels. For training, a random crop of 256 $\times$ 256 pixels is selected for the mini-batch.  The associated translational shift induced by cropping is recorded for the images containing a target.  For each image in the mini-batch, the network estimates the class and translational shift (when the image is of a target class) and the errors are backpropagated appropriately. One epoch of training consists of the number of mini-batches required to see each image of the training set once on average.

CNNs perform best when lots of training data is available.  In many situations though, large amounts of training data are not available. We call these instances \emph{low training data scenarios} and in them,  application of a domain prior becomes particularly important as its presence can significantly boost classification performance.  To study this effect, we trained each of the methods on a random (but consistent across methods) 10\% subset of the training.  \figurename \ref{fig:aucpr_as_function_of_training_pct} shows these results, and for convenience, the results when the full training data is available; recall these results are the AUCPRs of \figurename \ref{fig:precision_recall}. We show that the application of domain priors results in improved performance over all of the comparison methods when operating in low training data scenarios.  These results demonstrate how application of our domain priors, image enhancement and target localization, improve performance on both abundant and low training data scenarios. The priors we introduce utilize information  implicit during human interpretation and provide useful contextual information for our image classification method.


Deep networks often require a hyper-parameter search for optimal performance; SPDRDL is no different.  SPDRDL uses the RMSProp optimization scheme \cite{Tieleman2012} with a fixed learning rate of $10^{-4}$ which was determined through cross-validation.  Furthermore, the weights of the domain priors in Eq (\ref{eqn:ipsas_loss}) were also found through cross-validation giving the best results when $\lambda_1=10^{-6}$ and $\lambda_2=10^{-1}$.

\subsection{Ablation Study: Impact of Domain Priors}

Table \ref{table:ablation_losses} shows the performance of SPDRDL with no additional multi-task losses and the incremental addition of each domain prior. For the low and high training scenarios, each additional loss provides improved classification performance with both priors giving better performance than each individual prior.

Next, we compare SPDRDL against three common despeckling techniques to show the benefit of using the learned enhancement network with the SSP. Table \ref{table:ablation_losses_fixed_despeckler} shows the results for the high training scenario when we retrain the network by setting $\lambda_1=0$ in Eq (\ref{eqn:ipsas_loss}) and supplanting the enhancement network with one of the following despeckling filters: Gaussian filter, median filter, and total variation \cite{getreuer2012rudin, bush2011bregman}.  We can see that the best AUCPR performance of the pre-processed despeckled images is 0.9451 which is not as good as when the SSP is present, 0.9538.

\begin{table}[t]
    \caption{We evaluate performance of each domain prior in our loss function, Eq (\ref{eqn:ipsas_loss}), to demonstrate their utility. AUCPR is reported on the test set for the high (100\% of training data available) and low (10\% of training data available) training data scenarios. Note, the enhancement network is still present in the CL and CL+SSCP scenarios but the associated regularization loss is removed from the objective function.}
    \centering
    \begin{tabular}{c c c}
        \hline
        \textbf{Domain Priors} & \textbf{10\%} & \textbf{100\%} \\
        \hline \hline
        None, Only Classification Loss (CL) & 0.8742 & 0.9281 \\        
        CL + Structural Scene Context Prior (SSCP) & 0.8919 & 0.9503 \\
        CL + Structural Similarity Prior (SSP) & 0.8969 & 0.9456 \\
        Both, CL + SSCP + SSP & \textbf{0.9079} & \textbf{0.9538} \\
        \hline
    \end{tabular}
    \label{table:ablation_losses}
\end{table}

\begin{table}[t]
    \caption{We evaluate the performance of our SSP domain prior against several off-the-shelf despeckling methods to show SSP's utility. We do this by retraining the network but removing SSP's associated loss term in Eq (\ref{eqn:ipsas_loss}) and the Enhancement Network in \figurename \ref{fig:network} (SSCP is still included).  We feed to network three types of despeckled imagery and report AUCPR using 100\% of the available training data. We see that despeckling does give some performance gains over the CL configuration of \tablename \ref{table:ablation_losses}, but not as much as when the SSP is active (CL+SSP and CL+SSP+SSCP configurations of \tablename  \ref{table:ablation_losses}).  Recall, our method (CL+SSCP+SSP) yields an AUCPR of 0.9538 as shown in \tablename \ref{table:ablation_losses}.}
    \centering
    \begin{tabular}{c c}
        \hline
        \textbf{Despeckling Algorithm} & \textbf{100\%} \\
        \hline \hline
        CL + SSCP + Gaussian Filter & 0.9450 \\
        CL + SSCP + Median Filter & 0.9409 \\    
        CL + SSCP + Total Variation  & 0.9451 \\
        \hline
    \end{tabular}
    \label{table:ablation_losses_fixed_despeckler}
\end{table}

\subsection{Comparison Against State of the Art}
\label{sec:comparisons}
We demonstrate the efficacy of  SPDRDL by comparing against three state of the art deep learning methods for sonar ATR and two recent shallow-learning methods.  The deep learning methods were trained using Tensorflow 1.13.1 \cite{tensorflow2015-whitepaper}. The number of trainable parameters for each network is given in \tablename  \ref{table:num_free_params}. Development of the deep learning algorithms and their specific parameters follows.

\textbf{Emigh, et al. (IOA SAS/SAR 2018)} \cite{emigh2018supervised}. This network is based on the Resnet-18 architecture, does not rely on Imagenet pre-training, ingests dual-band SAS images, and is a binary classifier with output of a single scalar indicating target/non-target score. We modify the network to input a single-band SAS image and output four classes by using a softmax function after the last dense layer. We use the categorical cross-entropy loss when training the network as binary cross-entropy loss was originally specified by the authors and we adapted it to this multi-class scenario. We trained using the Adam optimizer \cite{kingma2014adam} with a learning rate of $10^{-3}$.  The paper mentions decreasing the learning rate when a loss plateau occurs but does not give details on its parameters. In lieu of this, we forgo the learning rate schedule and invoke the same early stopping rule used during SPDRDL training.

\textbf{Galusha, et al. (SPIE 2019)} \cite{galusha2019deep}. This network has only a few layers and is an Alexnet-like architecture. Like Emigh, et al., this network originally consumed dual-band SAS imagery and output a binary classification score representing target/non-target. We modify the network in a similar fashion as Emigh, et al. by using only a single-band SAS image as input and modify the output to support classification of four classes using a softmax activation after the last dense layer.  As in Emigh, et al., we use the categorical cross-entropy loss when training the network as the binary cross-entropy loss was originally specified by the authors. We train the network in the same fashion the authors used in their work: stochastic gradient descent for 2,000 epochs at a learning rate of $10^{-3}$.

\textbf{DensetNet121 (CVPR 2017)} \cite{huang2017densely}.  This is a common state-of-the-art off-the-shelf DenseNet architecture with 121 layers pre-trained on the Imagnet dataset. We choose this network for comparison because it is the feature extraction network used in SPDRDL, and serves as a baseline to provide evidence demonstrating our proposed priors improve classification performance. As with SPDRDL, use use the focal loss instead of the cross-entropy loss in order to demonstrate that our performance gains are not simply from this different classification loss. Densenet121 ingests three-channel color imagery imagery; we simply replicate the SAS input image over two additional channels to arrive at a three-channel image.  Finally, we use the global average pooling option after the feature extraction layer and apply a four class output with softmax. We train using the same procedure as SPDRDL.


\textbf{BoW-HOG (CVPR 2015)}. This method is a bag-of-words using histogram of oriented gradients features inspired by \cite{Isaacs_2015_CVPR_Workshops}. A comparison to a similar algorithm was also made by \cite{mckay2017robust}. For this approach, each image is divided into $16\times16$ pixel tiles of which the HOG features are computed. These features are then clustered using mini-batch k-means clustering \cite{sculley2010web} into what are known as \emph{words} in this setup.  The clusters of words form the \emph{vocabulary} for the bag-of-words model.
        
    
The size of the vocabulary, $k$, and the regularization parameters for the SVM, $C$, are chosen using a random search \cite{bergstra2012random, scikit-learn} of fifty iterations. The hyper-parameters for the search are chosen from $C\sim 10^{U[-5,10]}$ and $k\sim \round{U[10,500]}$ where $U$ represents the uniform distribution over the specified interval.  The hyper-parameter search returned best results for $c=4.36, k=402$.  A radial basis function kernel was used. BoW-HOG method is costly to compute so we use no data augmentation during training and measure performance solely on the center crops of Dataset B.

\textbf{DSRC (IEEE TGRS 2017)}. This method is a dictionary sparse reconstruction (DSRC) algorithm inspired by \cite{mckay2017robust}. We use mini-batch dictionary learning \cite{mairal2009online} with coordinate descent to learn the dictionary atoms.

At test time, the learned dictionary for each class is used to reconstruct the test images one class at a time.  The inverse of these errors from each reconstruction is transformed by the softmax function to class probabilities.  

The sparsity of the $L_1$ reconstruction loss,  $\alpha$, and the number of dictionary atoms per class, $k$, are chosen using a random search of fifty iterations.  The hyper-parameters for the search are chosen from $\alpha\sim 10^{U[-5,6]}$ and $k\sim \round{U[10,250]}$ and the best results were $\alpha=10^{-2.02}, k=147$.  A mini-batch size of ten was used for dictionary learning.

Similar to the BoW-HOG method, the extensive compute resources necessitated by this algorithm lead to using only center crops of the images of the training and validation sets.  As with the BoW-HOG method, performance is only reported on the validation set.

\begin{table}[t]
    \centering
    \caption{Number of trainable parameters for the deep learning methods.}
    \begin{tabular}{c c}
        \hline
        \textbf{Network} & \textbf{Number of Trainable Parameters} \\
        \hline \hline
        SPDRDL &  $\approx 9.16\times10^{6}$ \\
        Densenet121 &  $\approx 6.95\times10^{6}$ \\
        Emigh, et al. & $\approx 11.2\times10^{6}$  \\
        Galusha, et al. & $\approx 243\times10^{6}$\\
        \hline
    \end{tabular}
    \label{table:num_free_params}
\end{table}

\subsection{Results and Analysis}
In this section, we present results comparing SPDRDL against the several contemporary methods, demonstrate the necessity of each prior in our loss function formulation, analyze properties of the learned image enhancement function, and demonstrate the ability to reduce the network size considerably for use in low power embedded systems while maintain good classification performance.

\subsubsection{Classification Task}
\begin{figure*}[t]
    \begin{tabular}{c c c}
        \includegraphics[width=0.3\linewidth]{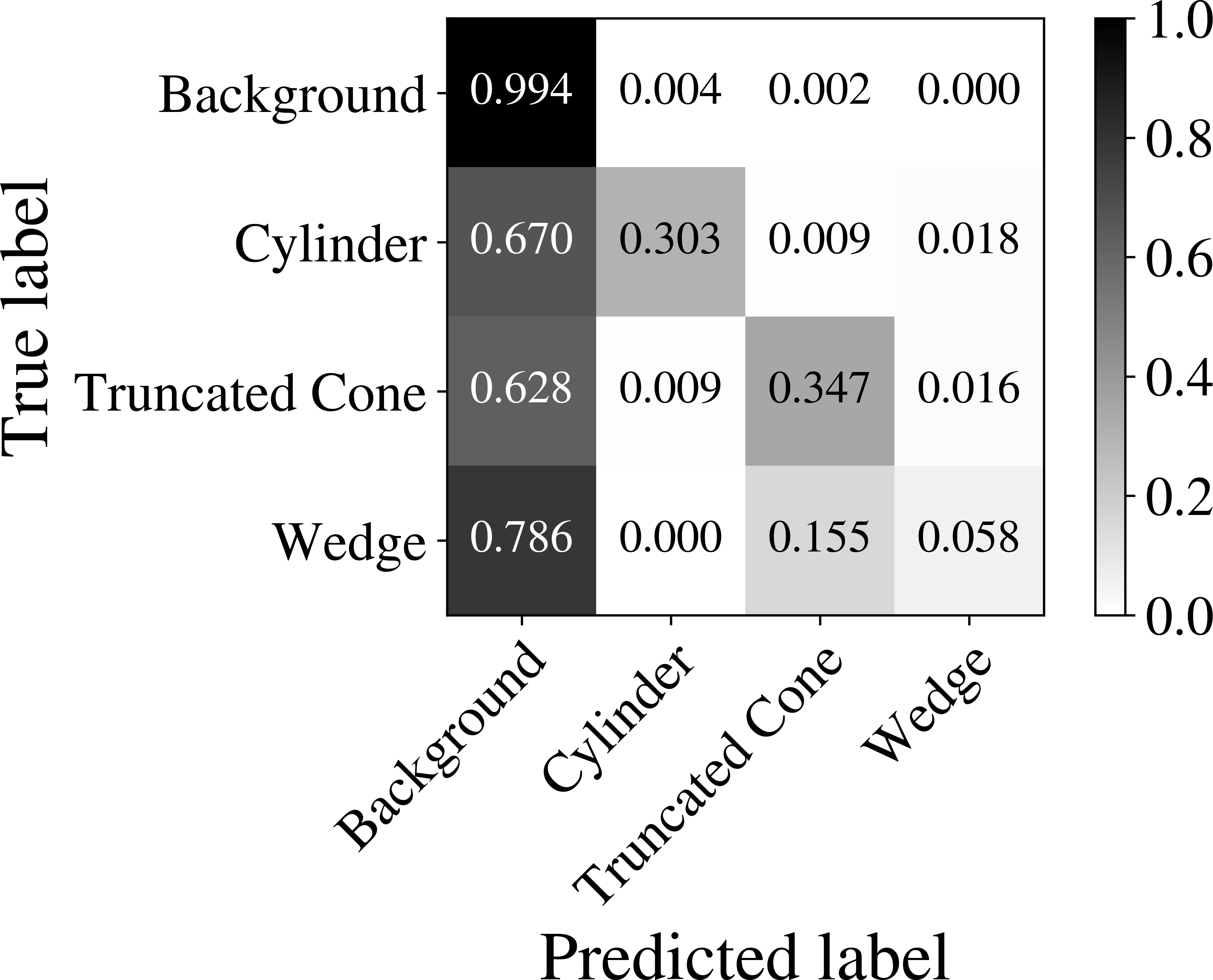} &
        \includegraphics[width=0.3\linewidth]{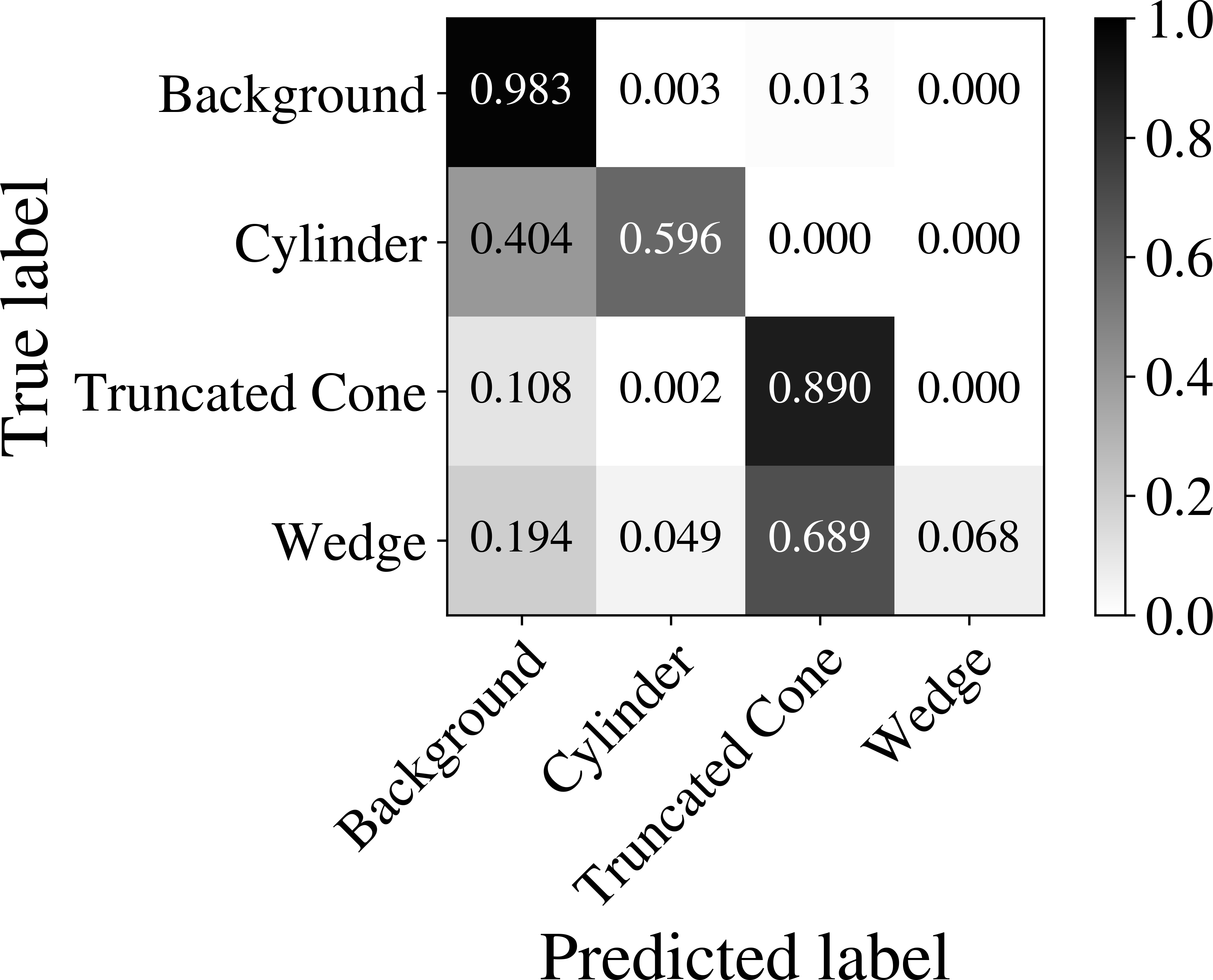} &
        \includegraphics[width=0.3\linewidth]{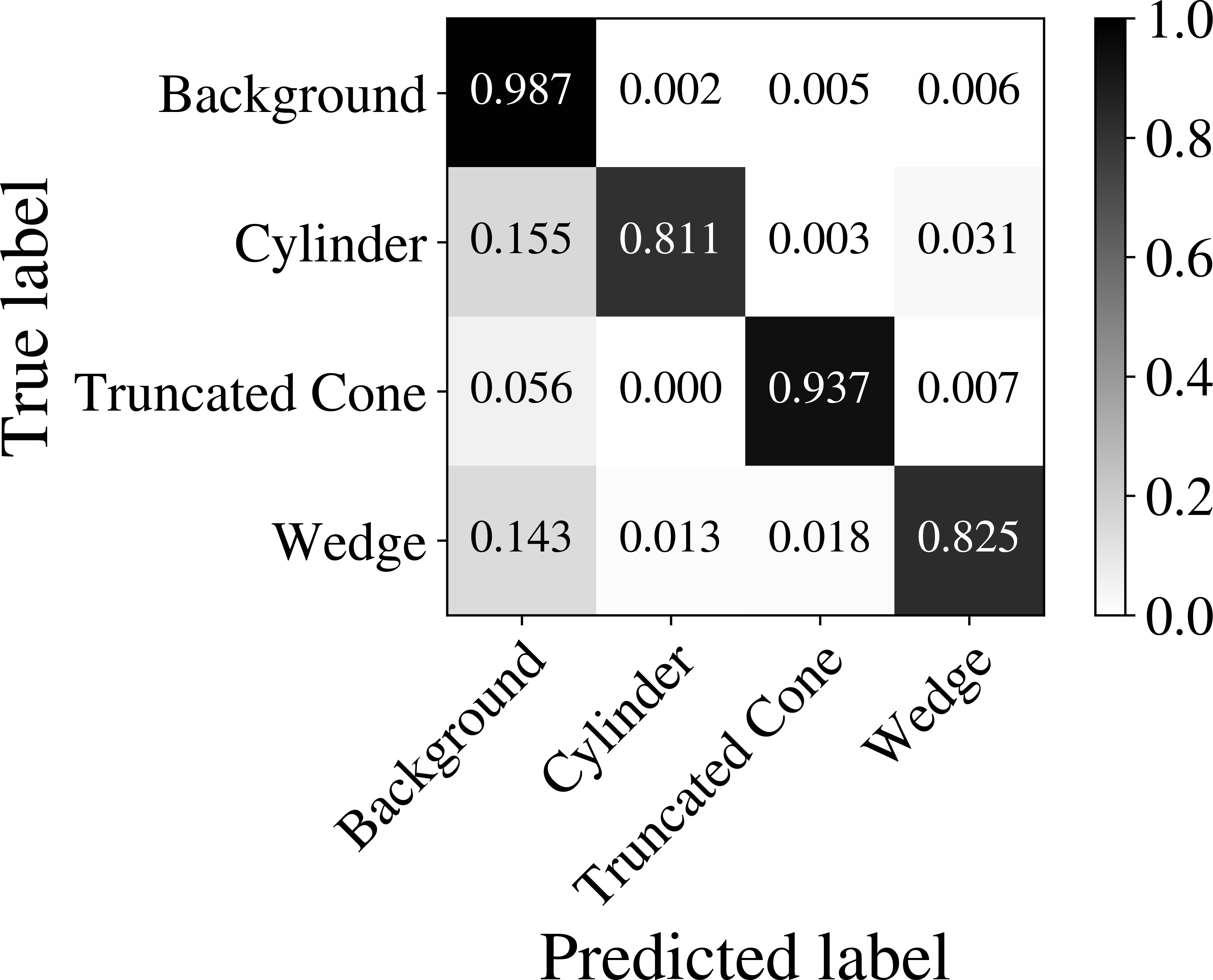} \\
        (a) BoW-HOG & (b) DSRC & (c) Emigh, et al. \\
        & & \\
        
        \includegraphics[width=0.3\linewidth]{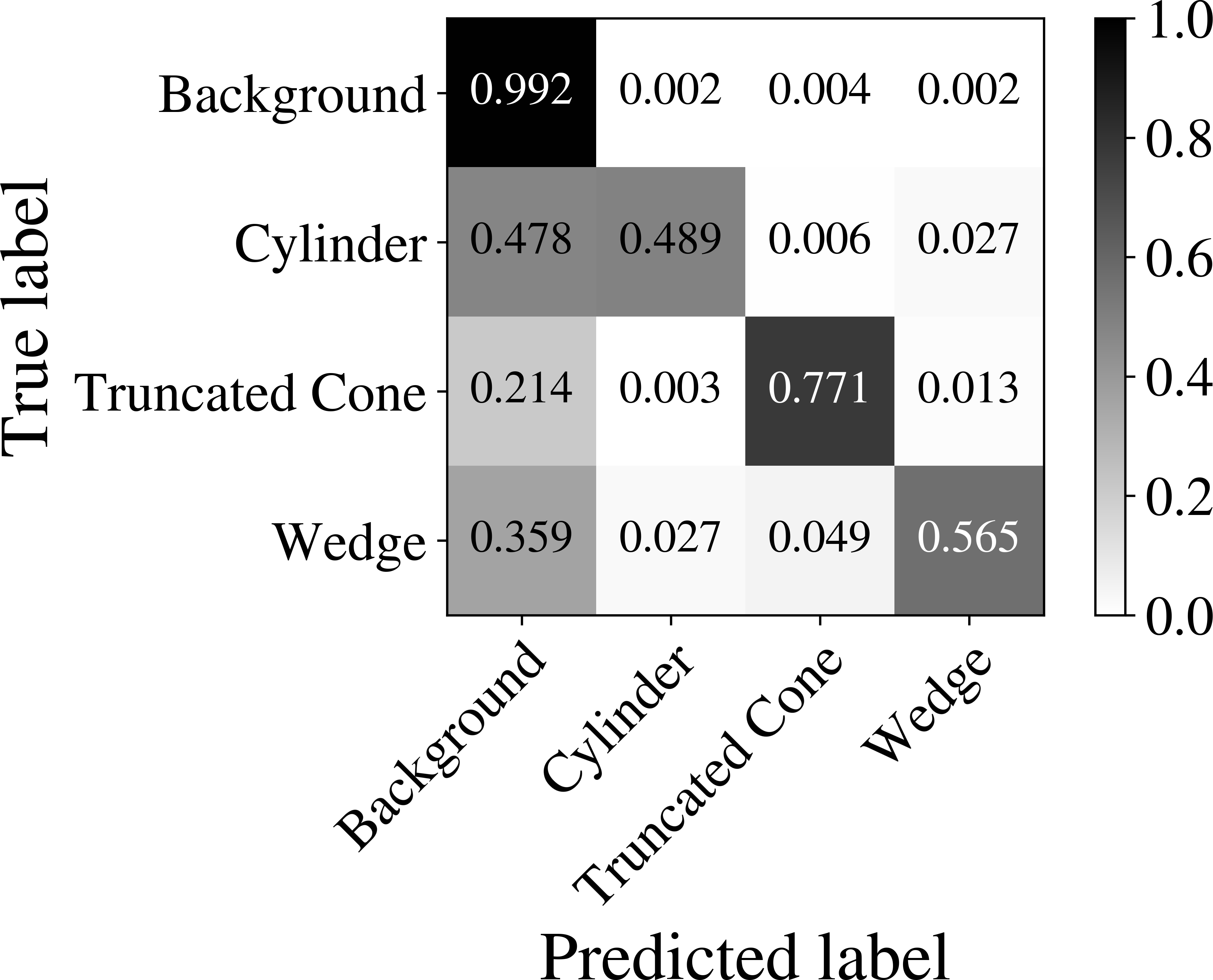} &
        \includegraphics[width=0.3\linewidth]{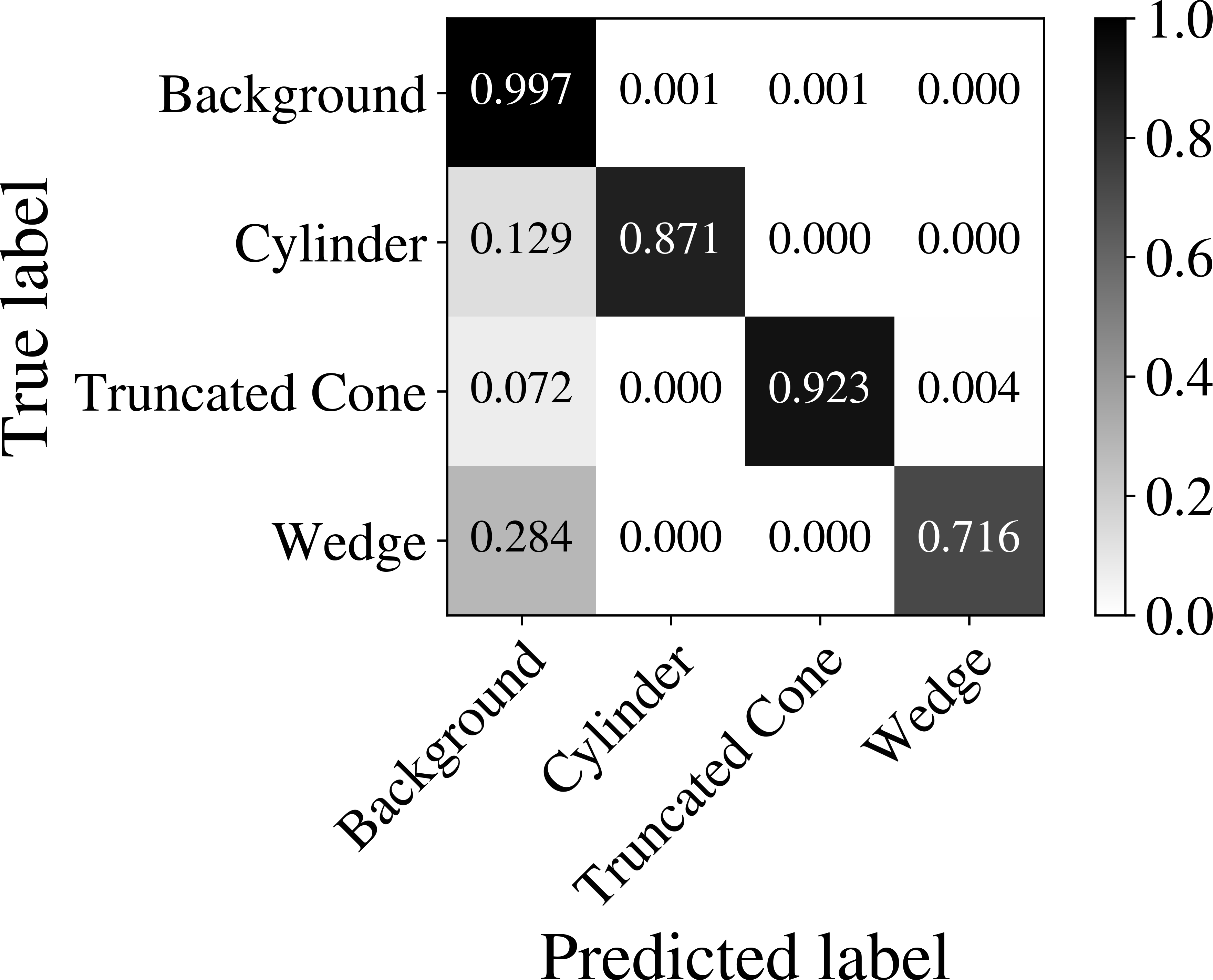} &
        \includegraphics[width=0.3\linewidth]{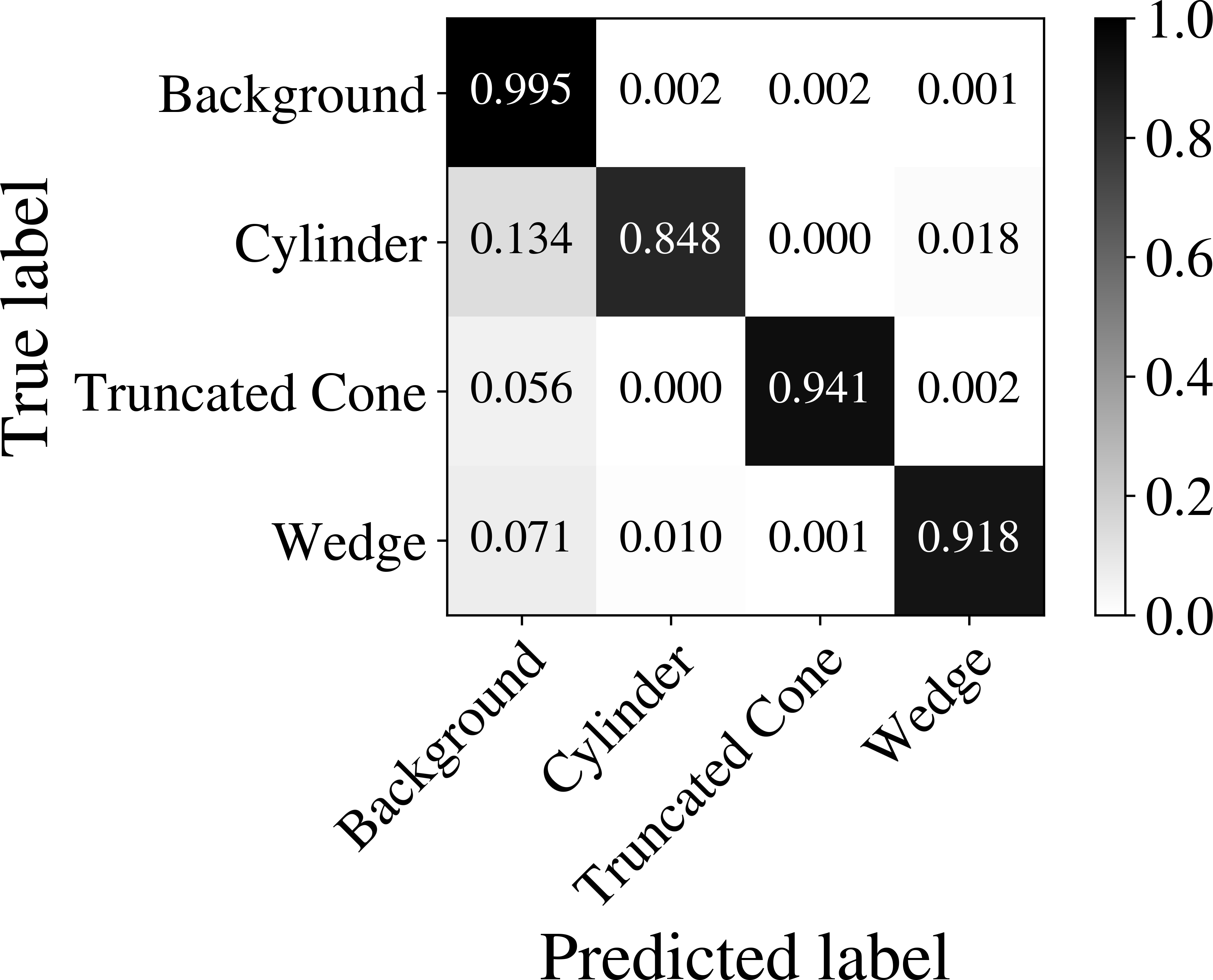} \\      
        (d) Galusha, et al. & (e) Densenet121 & (f) SPDRDL  
    \end{tabular}
    \caption{Confusion matrices for SPDRDL and comparison methods. Larger numbers along the diagonal indicate better performance. We see the benefits of the added domain priors through improved performance, especially of the wedge class.}
    \label{fig:confusion_matricies}
\end{figure*}

\begin{figure*}[t]
    \begin{tabular}{c c c}
        \includegraphics[width=0.3\linewidth]{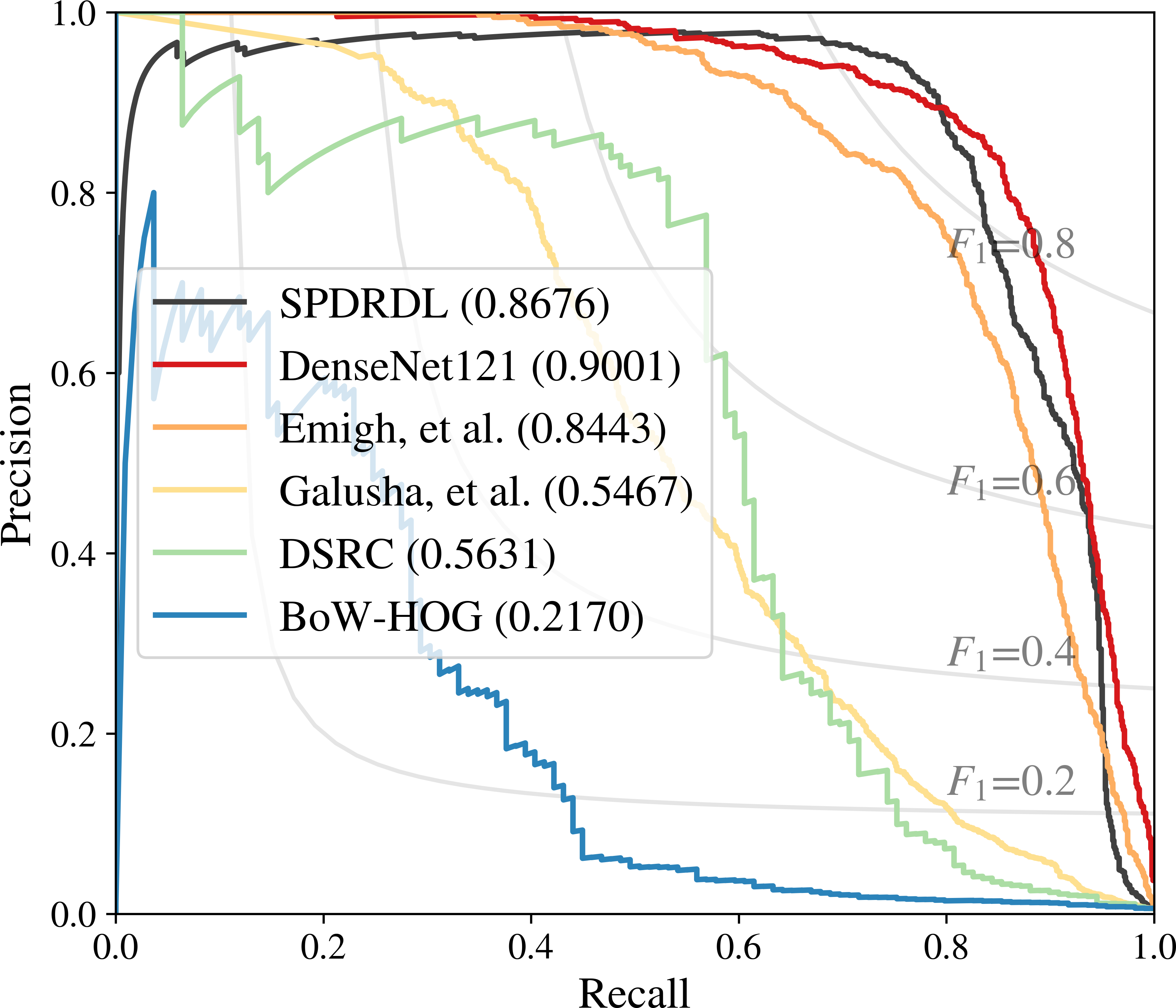} &
        \includegraphics[width=0.3\linewidth]{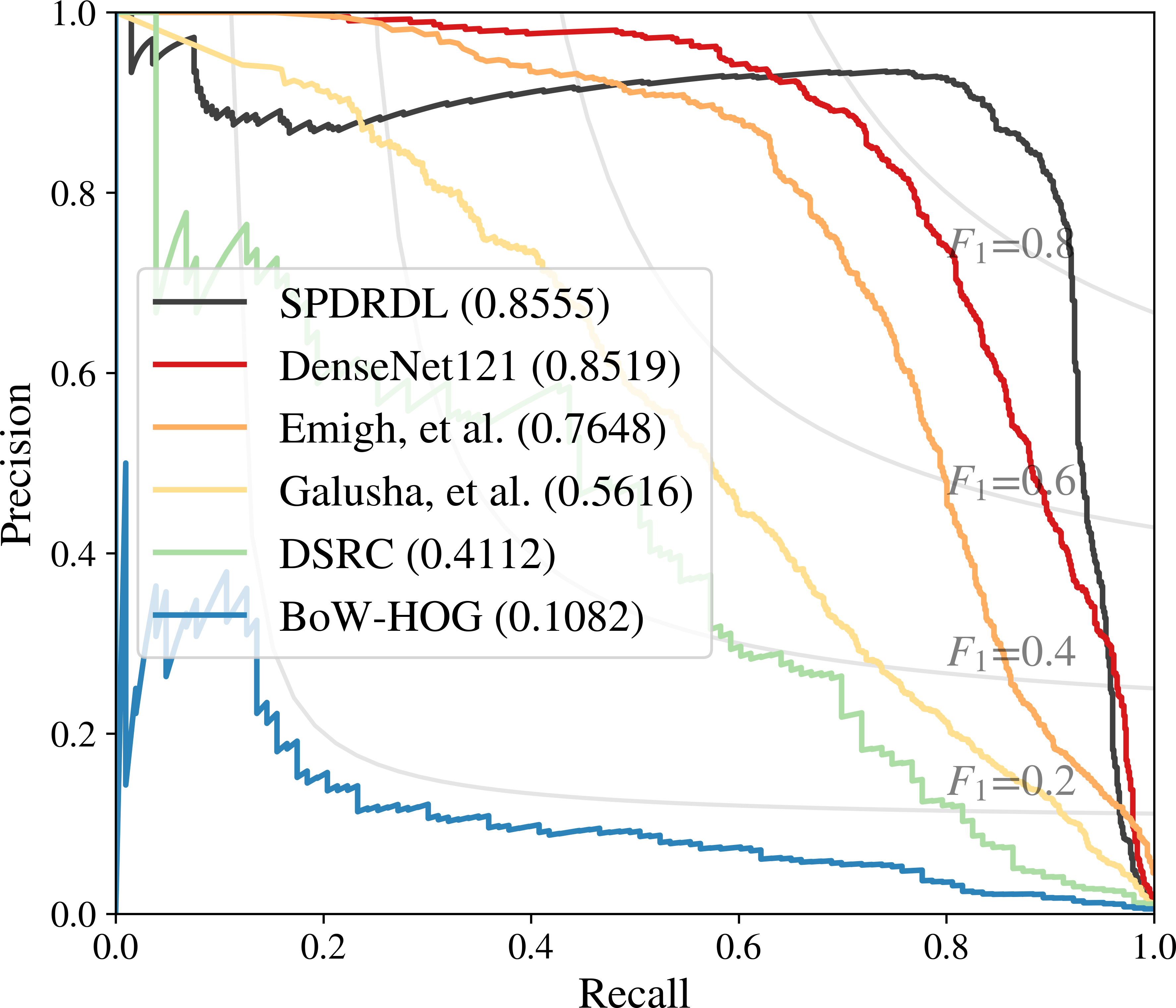} &
        \includegraphics[width=0.3\linewidth]{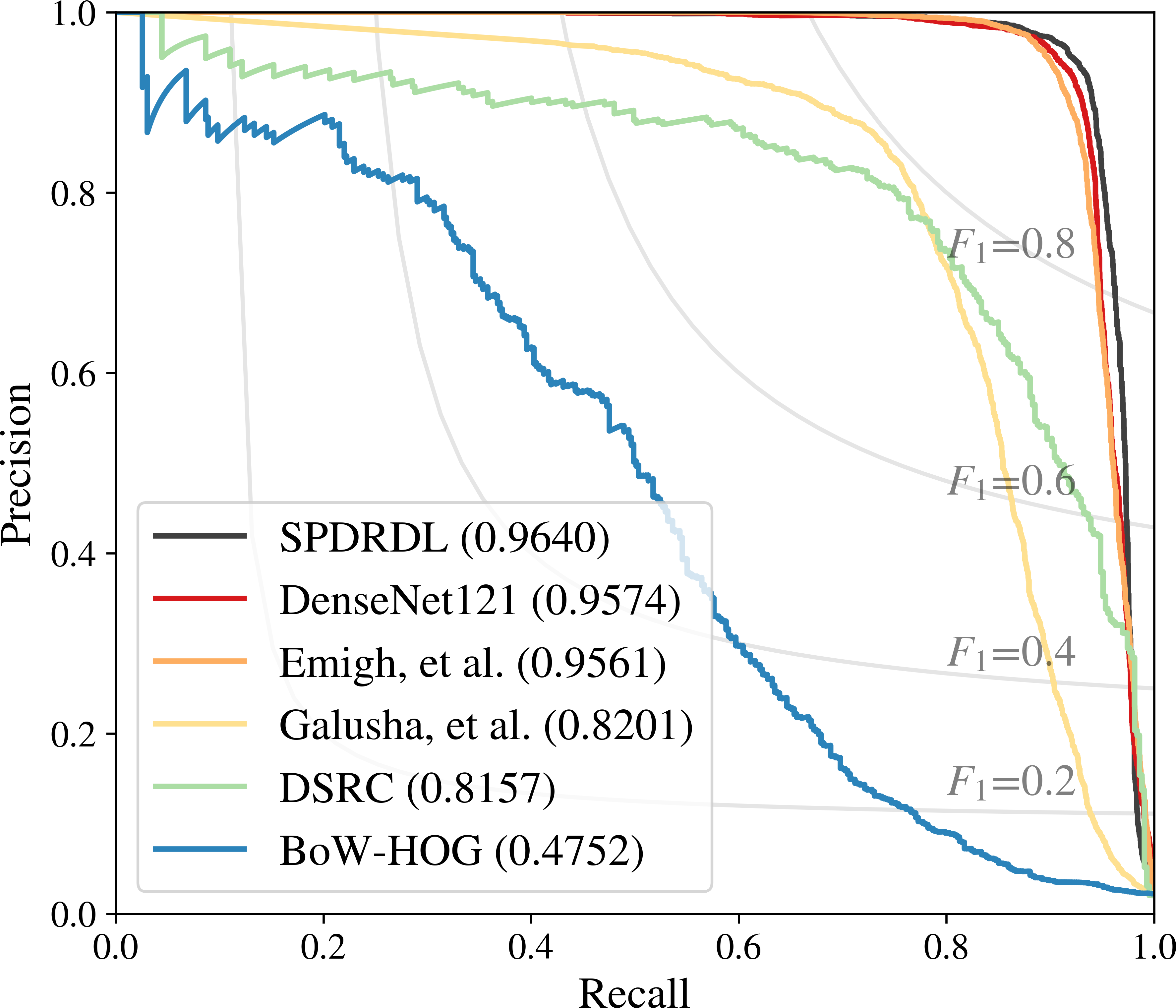} \\
        (a) Cylinder & (b) Wedge & (c) Truncated Cone
    \end{tabular}
    \caption{Precision-recall curves for target type: (a) cylinder, (b) wedge, (c)  truncated cone. AUCPR in parentheses; larger values are better.}
    \label{fig:precision_recall_targets}
\end{figure*}

\begin{figure}[t]
    \centering
    \includegraphics[width=0.8\linewidth]{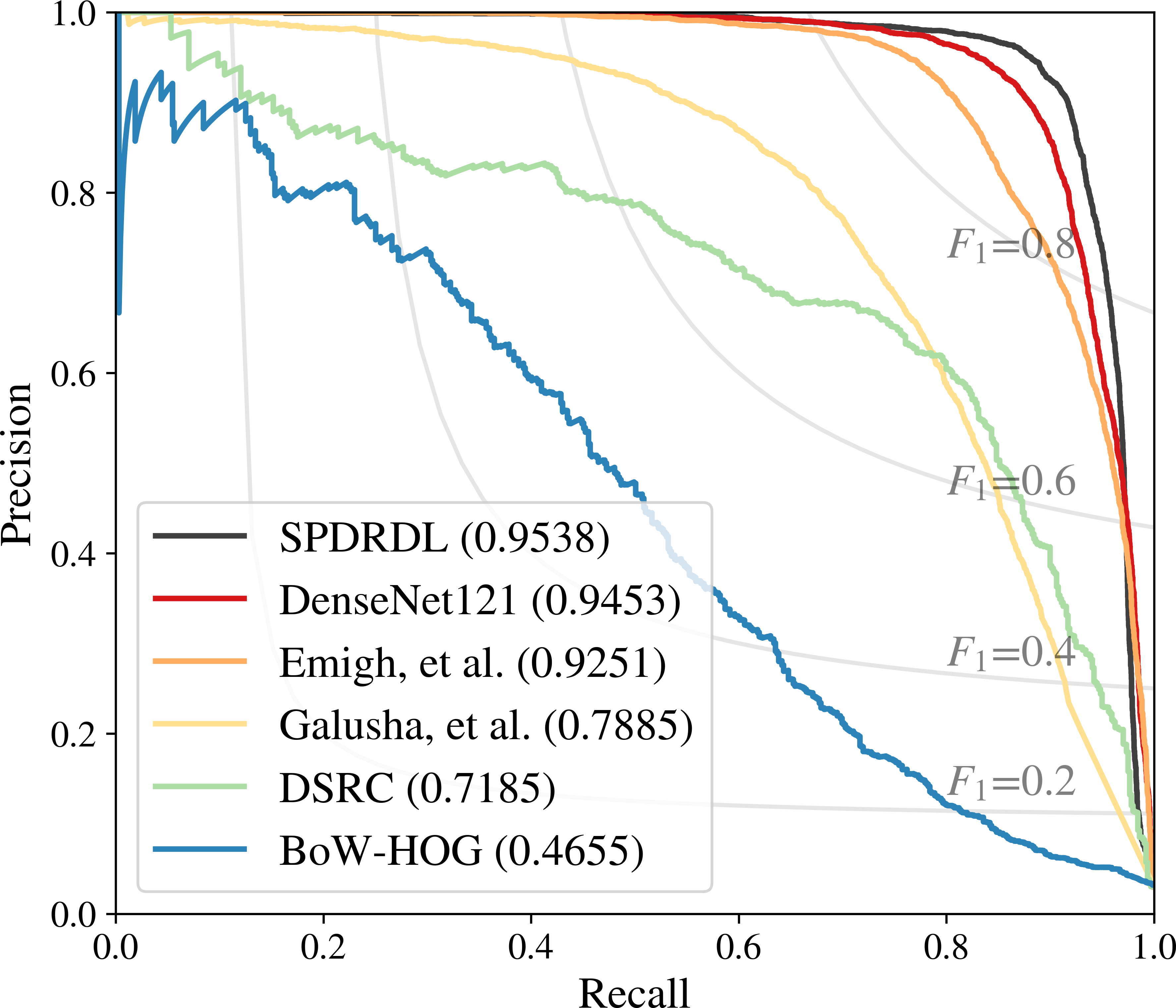}
    \caption{PR curves for all methods obtained using a one-versus-all method. AUCPR in parentheses; larger values are better.}
    \label{fig:precision_recall}
\end{figure}

We show confusion matrices in \figurename \ref{fig:confusion_matricies}, precision recall curves for each target class versus background in \figurename \ref{fig:precision_recall_targets}, and precision recall curves for target versus background in \figurename \ref{fig:precision_recall}. We demonstrate the learning efficiency of our method by showing results from the ablation study.  From these figures, we can see the benefits our domain priors afford us.  In almost all metrics, SPDRDL outperforms existing methods.

\begin{figure}[t]
    \centering
    \includegraphics[width=0.8\linewidth]{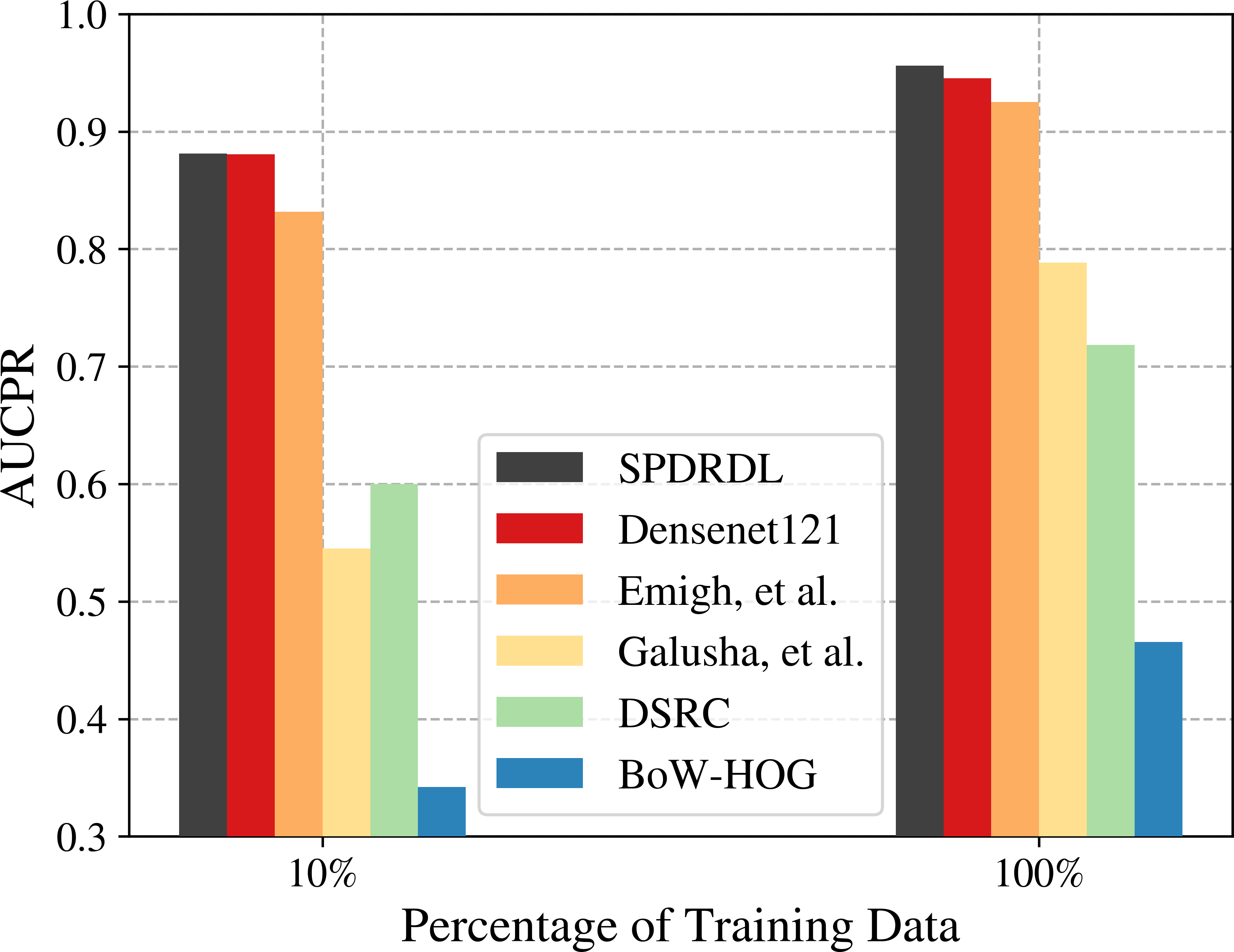}
    \caption{Deep learning methods work well when the training data is plentiful.  Sometimes such a situation is not possible as collection of training data can be expensive. To demonstrate the necessity of the domain priors our method introduces, we train each method using all the training data available and then train using a random 10\% subset.  We can see from the low training data scenario that SPDRDL still produces quite competitive results indicating the increased performance granted by the domain priors.}
    \label{fig:aucpr_as_function_of_training_pct}
\end{figure}

\begin{figure}[t]
    \centering
    \includegraphics[width=0.8\linewidth]{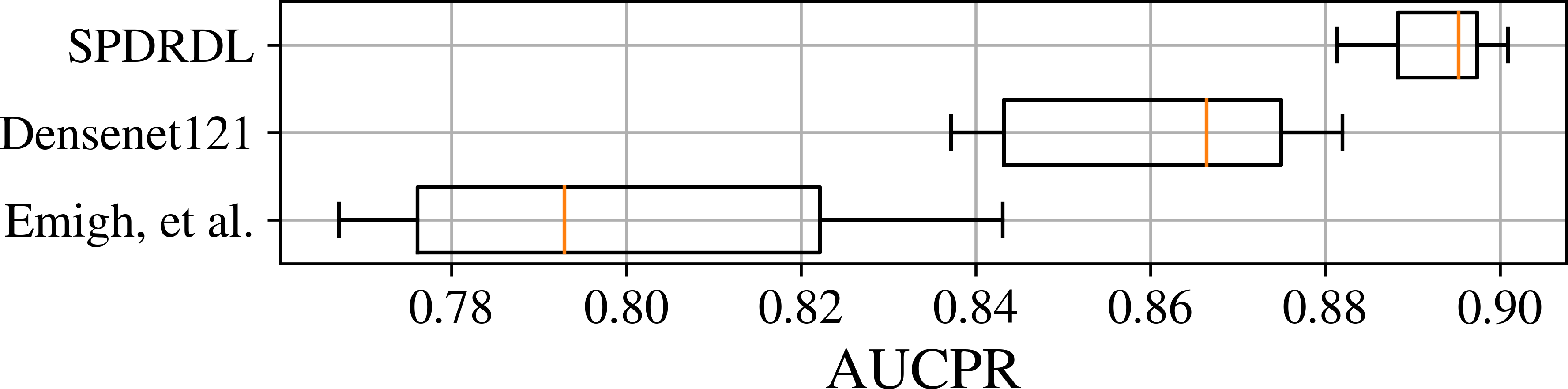}
    \caption{To understand the sensitivity of selection bias in a low training data scenario, we trained the top three methods each ten times with a random sample of 10\% of the training data and plot the results. Due to the large run-time associated with evaluating all the test set imagery, we report results on the validation set only. Larger AUCPR indicates better performance. As shown, we can clearly see a benefit in performance with the addition of the domain priors over the next best method, Densetnet121. }
    \label{fig:sensitivity_analysis}
\end{figure}

Indeed, the results of \figurename \ref{fig:aucpr_as_function_of_training_pct} show SPDRDL performance very similar to Densenet121 in the low training data scenario with SPDRDL exhibiting a slight performance gain.  However, the gap between SPDRDL and Denset121 may be much larger in reality. In this case, it is quite likely we are seeing the effects of selection bias of the training data subset.  To examine issues of sample selection bias, we train the top three methods ten times each using a random subset of 10\% of the training data; the same subsets were used for each algorithm evaluation.  Evaluating the test set on all of these trained models would be computational prohibitive due to the large test set size.  Therefore, we report on the validation set performance for the best epoch of each. As shown in \figurename \ref{fig:sensitivity_analysis}, the performance gap between SPDRDL and the next best method, Densenet121, becomes much more distinct when viewing the results in the context of selection bias. 

We also examined the performance gain of each regularization term of Eq (\ref{eqn:ipsas_loss}) to demonstrate its necessity. This was done by setting each loss term to zero and then retraining and reevaluating the network. Consequently, we can see both priors of Eq (\ref{eqn:ipsas_loss}) improve classification performance even in the low training data scenario demonstrating the benefits of incorporating domain knowledge.

\begin{figure}[t]
    \begin{tabular}{c c}

        \includegraphics[width=0.45\linewidth]{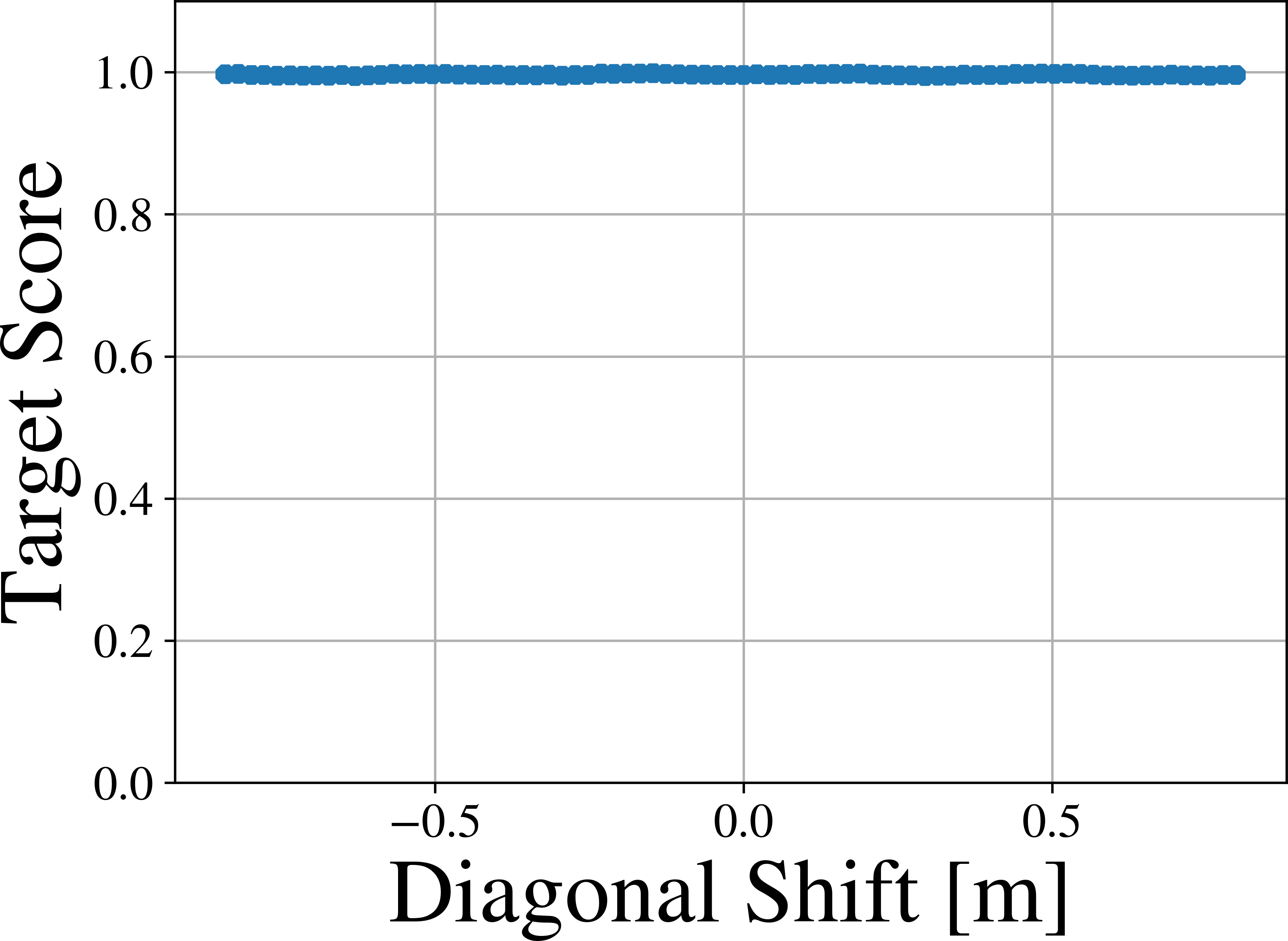} &
        \includegraphics[width=0.45\linewidth]{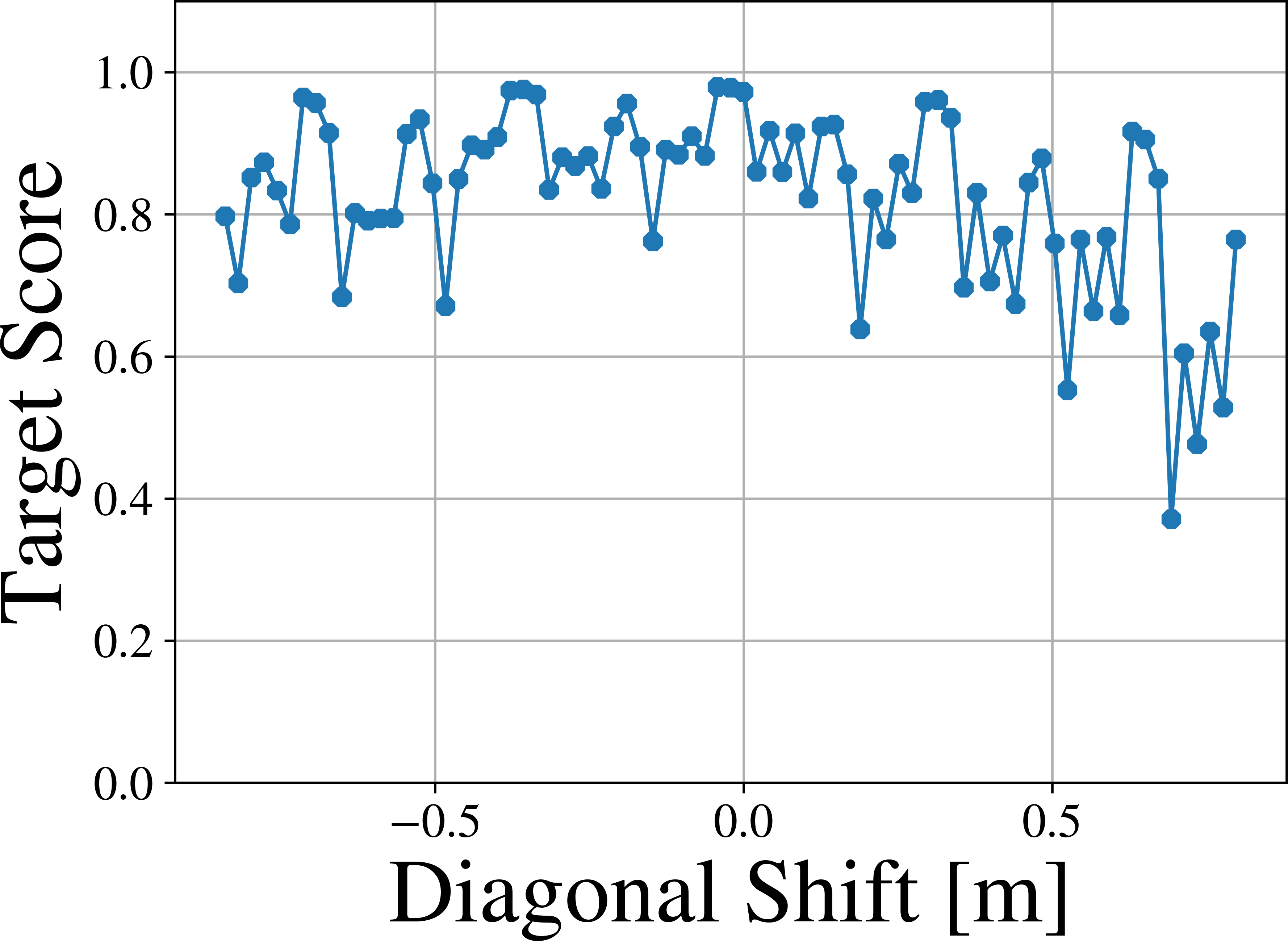}  \\
        (a) SPDRDL & (b) Densenet121
     \end{tabular}

     \begin{tabular}{P{\linewidth}}
     \includegraphics[width=0.35\linewidth]{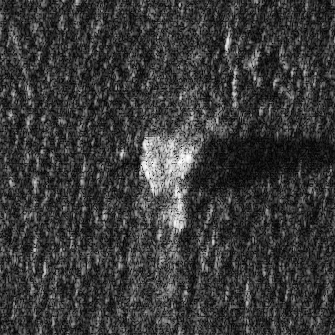} \\
     (c) Image under test for (a) and (b) which contains a target-class object.
     \end{tabular}
    \caption{(a) and (b) Target scores (SPDRDL/Densenet121 predicted classification probability) as a function of diagonal image translation of the top two performing algorithms for a sample input image. We can see for the Densenet121 method that a small translation of the input image results in an unpredictable classification whereas SPDRDL does not exhibit this. (c) Image under test for (a) and (b). 
    }
    \label{fig:translation_invariance_demo}
\end{figure}

\begin{table}[h]
    \centering
    \caption{Classifier scores were compiled for all nine shifts of each image in the test set. Next, we compute the standard deviation of scores for each images.  Finally, the table shows the mean of those scores over the entire test set; lower number indicate more shift invariance (i.e. better performance). We see that SPDRDL has the best translation invariance of all the methods.  We hypothesize that Densetnet121 has the second best performance because it contains only a single instance of MaxPooling and has many levels of feature averaging.}
    \begin{tabular}{c c}
        \hline
        \textbf{Network} & \textbf{Shift Invariance Score (Lower is Better)} \\
        \hline \hline
        SPDRDL & {$\mathbf{5.41\times 10^{-4}}$} \\
        Densenet121 & $5.82 \times 10^{-4}$ \\ 
        Emigh, et al. & $2.17 \times 10^{-3}$ \\
        Galusha, et al. & $3.94 \times 10^{-2}$ \\
        \hline
    \end{tabular}

    \label{table:translation_invariance_performance}
\end{table}

We can see from the results that the SPDRDL method outperforms the deep learning methods and the shallow learning methods by a significant amount demonstrating the usefulness of the domain priors. 
For our ablation analysis, we focus on two aspects: (1) Generalization efficacy using in a low training data scenario where only 10\% of the training data is used. (2) Necessity of the additional loss terms from our domain priors used in Eq (\ref{eqn:ipsas_loss}).

\figurename \ref{fig:aucpr_as_function_of_training_pct} shows AUCPR for all methods using 10\% and 100\% of the training data; SPDRDL outperforms all the comparison methods showing the efficacy of the additional domain priors. 

\begin{figure}[t]
    \centering
    \includegraphics[width=0.9\linewidth]{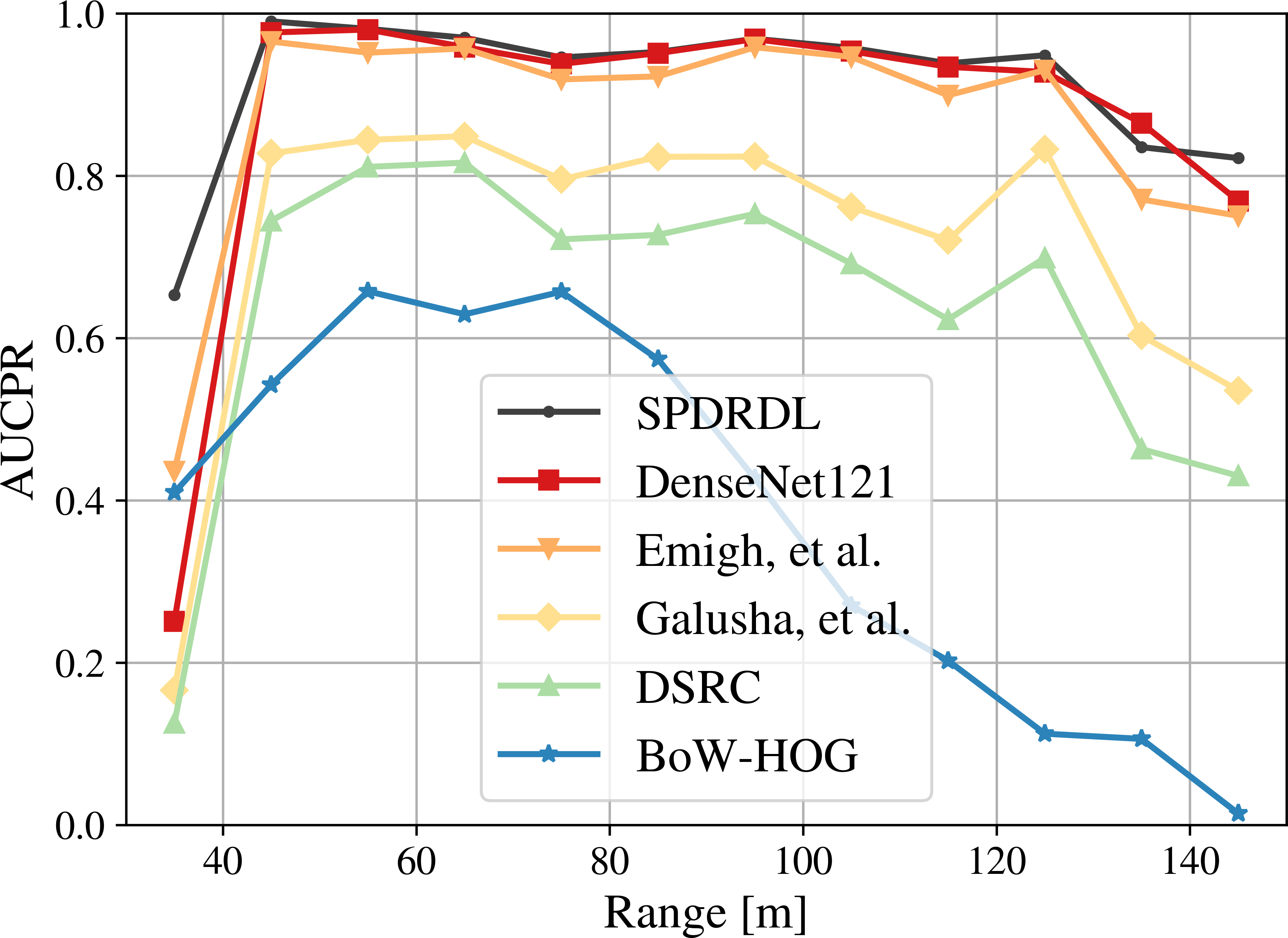} \\
    \includegraphics[width=0.9\linewidth]{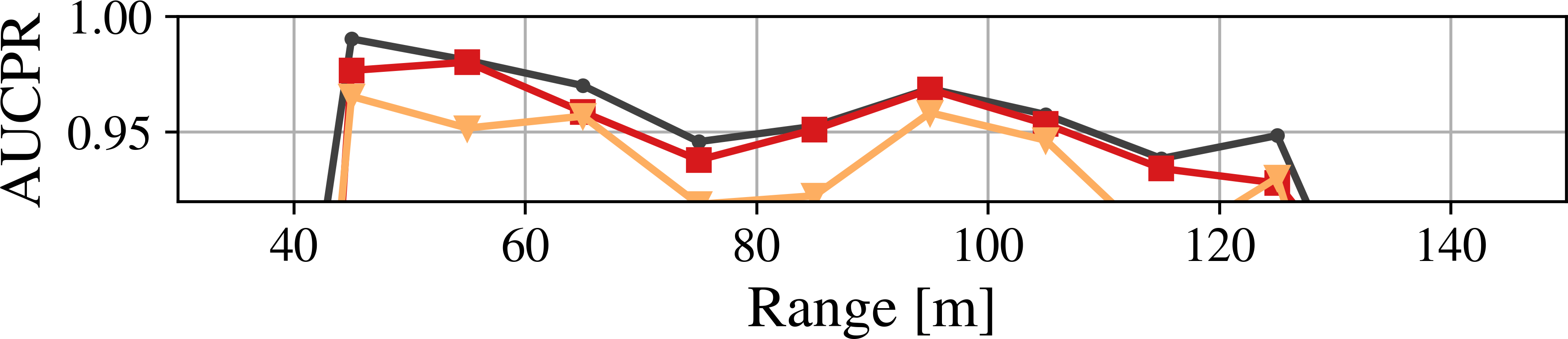}
    \caption{Top panel, AUCPR as a function of range from the sonar for all methods. Larger values indicated better performance. Bottom panel is a zoom of the top panel highlighting the differences of the top three methods: SPDRDL, DenseNet121, and Emigh, et al.}
    \label{fig:aucpr_with_range}
\end{figure}

Finally, we examine the two-class performance as a function of object range from sensor.  \figurename \ref{fig:aucpr_with_range} shows SPDRDL performing well over an extensive range from the sonar due to the addition of the domain priors. Especially at the nearest and furthers ranges from the SAS, SPDRDL outperforms the comparison methods.

\subsubsection{Image Enhancement Task}
Because of speckle noise, image despeckling algorithms are often employed to enhance SAS images for the purpose of improving human interpretability. We posit that there exists an image enhancement function which improves classification performance and we call this the structural similarity prior; any enhancement algorithm must preserve the structural similarity between the input and output images. We employ this prior in SPDRDL through the use of a learned image enhancement transform which is constrained by the loss of Eq (\ref{eqn:ms_ssim}).

\begin{figure}[t]
    \centering
    \begin{tabular}{c c}
        \includegraphics[width=0.47\linewidth]{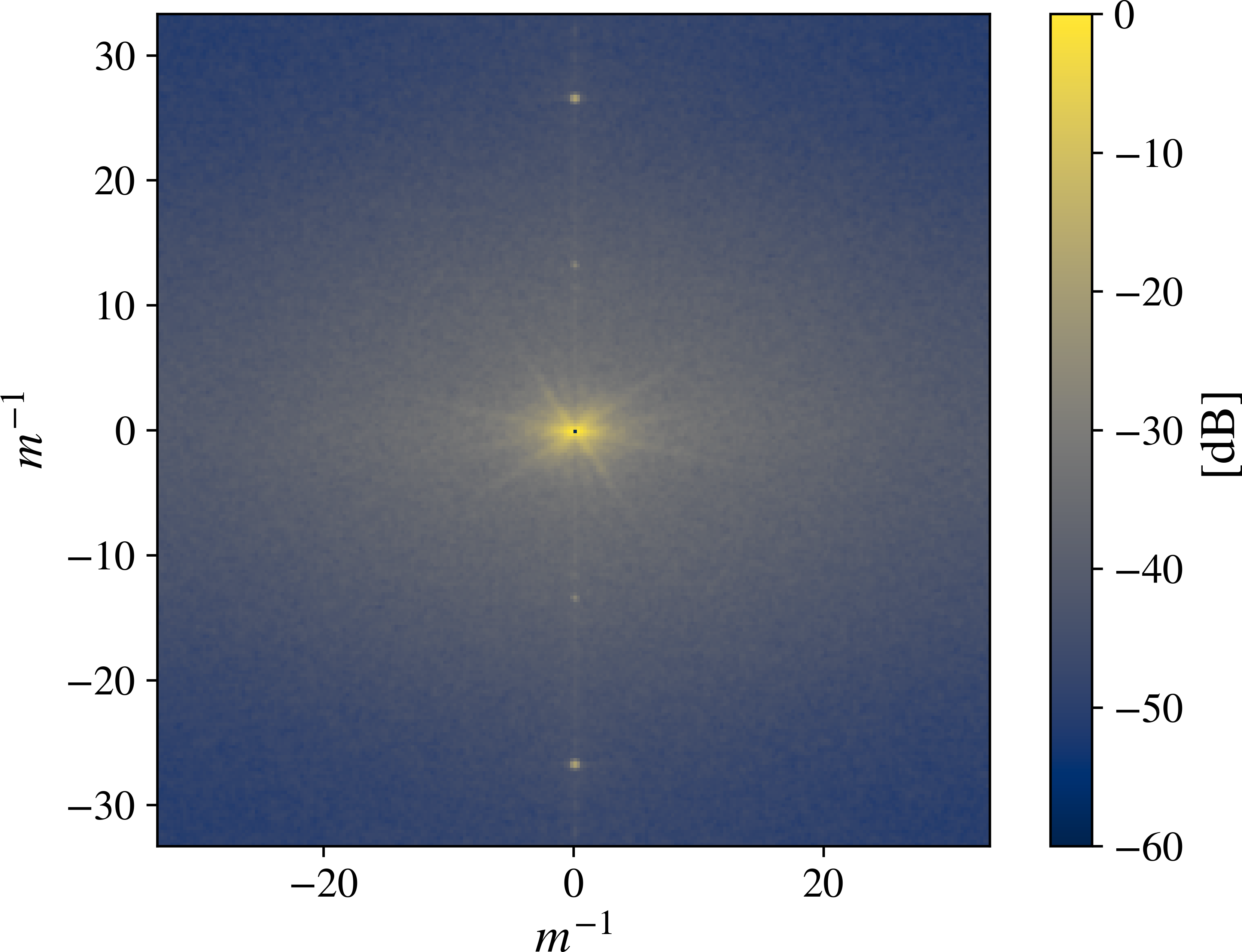} &     \includegraphics[width=0.47\linewidth]{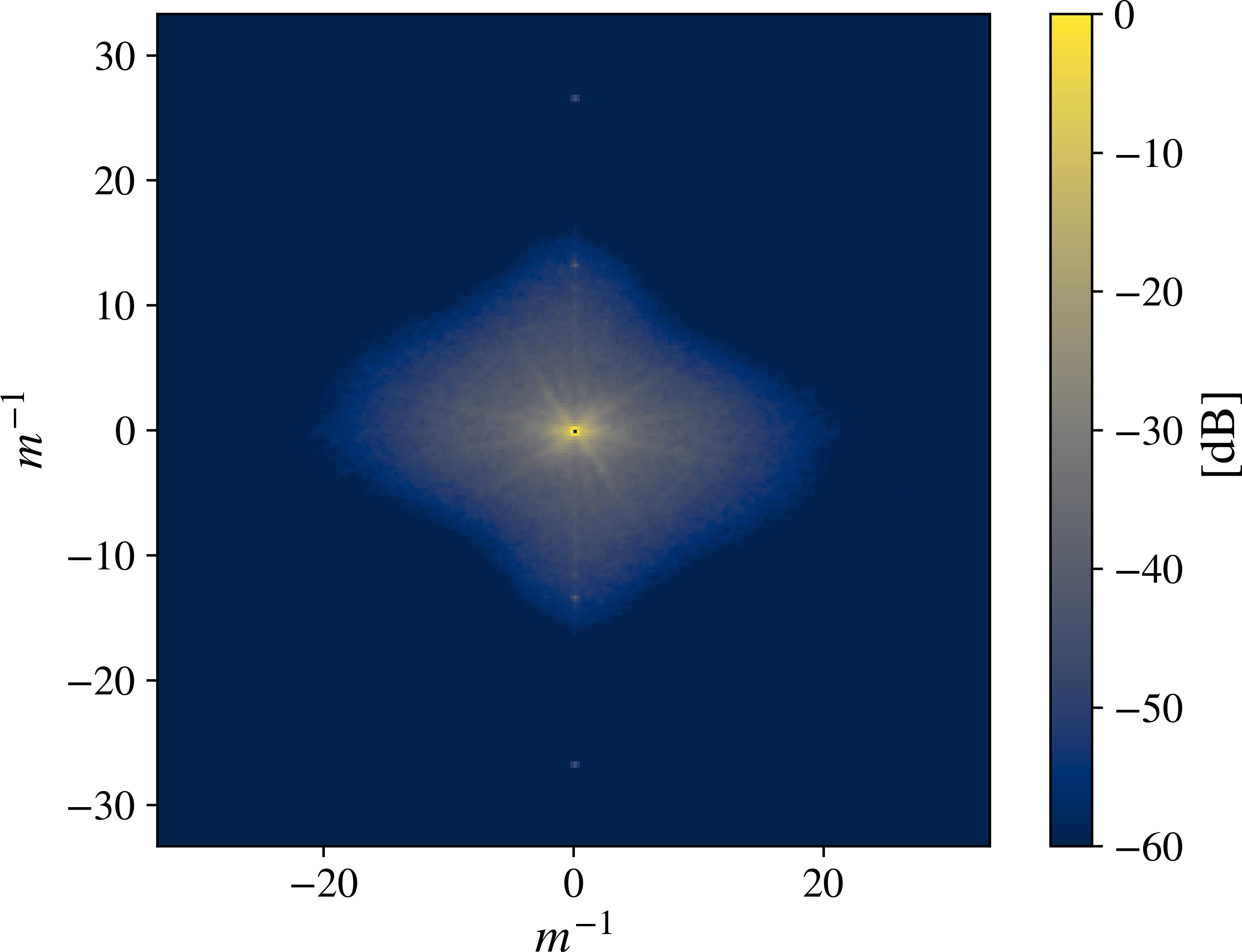} \\
        (a)  & (b) 
    \end{tabular}
    \caption{We analyze the frequency spectrum of the original images input to the network (a) and of our learned image enhancement (b).  In panel (b), we can see selective attenuation across spatial frequencies and orientations due to the use of the data-adaptive image enhancement prior.}
    \label{fig:image_spectra}
\end{figure}

We analyze the output of the enhancement network by examining its frequency response and compare it to the frequency response of the original images. That is, we compute the frequency response all the images in target classes and present their averaged spectra; all images are windowed with a 2D hamming window prior to frequency domain conversion. \figurename \ref{fig:image_spectra} depicts these results.  We can see selective suppression in both spatial frequency and orientation from the integration of our image enhancement network. This behavior is in contrast to what one would get with a simple 2D Gaussian filter which would give an isotropic frequency response.


\subsubsection{Target Localization Task and Translation Equivariance}
CNNs lose their translation equivariance through the addition of non-unitary strided pooling and convolution layers. In this section, we quantify this phenomenon over the deep learning methods to demonstrate the necessity of adding anti-aliasing filters before subsampling in the network.  To first illustrate the problem, we show classification scores of an image over a large set of translations, \figurename \ref{fig:translation_invariance_demo}.  As shown, classification scores can vary drastically even at small pixel shifts. For example, suppose a well-centered target is presented to the classifier and is correctly classified.  By translating the image by one pixel, the classifier will now misclassify it!

We analyze the translation invariance performance of each deep learning algorithm to assess its ability in providing translation invariance.  We only examined the target/background scenario by converting the class prediction estimates to a single scalar from $[0,1]$ indicating target score. For a single image, we compute its scores for a center crop and eight extreme crops.  We then measure the standard deviation of these scores and call this metric, $\Psi$, the image's \emph{shift invariance score}. More formally, we define the shift invariance score as \begin{multline}
\Psi(f, \mathbf{x}) = \text{stdev}  ( \{f(\mathbf{x}_{-h,-w}), f(\mathbf{x}_{-h,0}), f(\mathbf{x}_{-h,w}), \\ f(\mathbf{x}_{0,-w}), f(\mathbf{x}_{0,0}), f(\mathbf{x}_{0,w}), f(\mathbf{x}_{h,-w}), f(\mathbf{x}_{h,0}),  f(\mathbf{x}_{h,w}) \} )
\end{multline}
where function $f$ is the inference model, $\mathbf{x}_{\text{a}, \text{b}}$ is the input image translated by $a$ and $b$, and  $h=w=59\text{cm}$. As a result of this formulation, lower shift invariance scores indicate better translation robustness. We report the results in \tablename \ref{table:translation_invariance_performance}. We can see our proposed method has the greatest translation invariance due to the addition of the anti-aliasing filters layers employing non-unitary stride. 

\subsubsection{Network Compute Burden and Reduction}
Deep networks consisting of a large number of parameters can be challenging to deploy on embedded hardware because of the large memory and computational footprint required.  However, we can reduce the number of parameters used during inference by utilizing a network pruning algorithm. Although such algorithms remove weights,  classifier performance is often maintained and in some cases, even improved. 

\begin{figure}[t]
    \centering
    \includegraphics[width=0.9\linewidth]{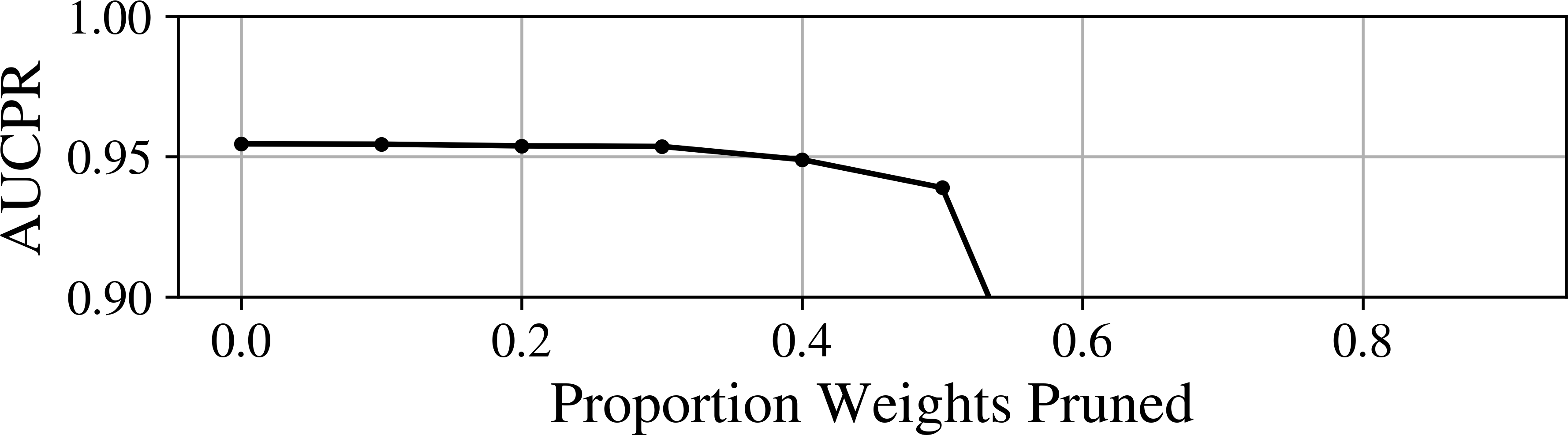}
    \caption{AUCPR as a function of weight pruning proportion.  Weights are sorted by magnitude and removed starting with the lowest magnitude weights first.  Significant pruning is accomplished while still maintaining good classification performance.}
    \label{fig:prune_results}
\end{figure}

We reduce the number of free parameters of network using the pruning method of \cite{frankle2018lottery}.  In this method, we sort the absolute value of the network weights and set the lowest proportion of weights to zero.  \figurename \ref{fig:prune_results} illustrates the results of performing this operation on SPDRDL.  We can see that even when the number of parameters is reduced by half, the network still results in competitive classification performance. 

\section{Conclusion} \label{section:conclusion}
SAS is relatively new field of remote sensing and has shown increased interest in the last two decades because of the high-quality imagery such systems produce.  Recent, technological advancements in computing (e.g. GPUs) have made it easy to deploy SAS-equipped UUVs with on-board image formation.  This capability paves the way for automatic machine interpretation of the imagery with the goal of influencing vehicle behavior.  Despite the success of SAS, high false alarm rates provide difficulty for autonomy to make decisions in situ.

In this work, we developed a SAS ATR algorithm exhibiting improved performance over state-of-the-art methods by integrating domain knowledge of SAS images previously overlooked by existing methods.  Our formulation jointly learns image enhancement and target localization for the purposes of improving the downstream task of image classification. We compare our method to several state-of-the-art techniques, including two recent deep-learning methods, and demonstrates its efficacy. Finally, we use a recently proposed pruning technique to show we can halve the number of free parameters in our network and still achieve competitive performance, thus demonstrating feasibility for real-time deployment.

Future work includes extending the approach to phase and frequency representations of the SAS images.
\section*{Acknowledgments}
The authors would like to thank the NATO Centre for Maritime Research \& Experimentation (CMRE) for providing the data used in this work. The collection of the data was funded by the NATO Allied Command Transformation. Colors used in the plots are derived from \cite{colorbrewer} and \cite{nunez2018optimizing}. I.G. would like to thank Dr. David Williams and Dr. John McKay for their helpful comments during the progress of this work.

\bibliographystyle{IEEEtran}
\bibliography{IEEEabrv,ref_tgars}


\end{document}